\documentclass[lettersize,journal]{IEEEtran}
\usepackage{amsmath,amsfonts}
\usepackage{algorithmic}
\usepackage{algorithm}
\usepackage{array}
\usepackage[caption=false,font=footnotesize]{subfig}
\usepackage{textcomp}
\usepackage{stfloats}
\usepackage{url}
\usepackage{verbatim}
\usepackage{graphicx}
\usepackage{cite}

\usepackage{multirow}
\usepackage{CJKutf8}
\usepackage{amsmath,amssymb,amsthm}
\usepackage{makecell}
\usepackage[table]{xcolor}
\usepackage{tabularx}
\usepackage{array}
\usepackage{multirow}
\begin{document}

\title{GJDNet: Robust Graph Neural Networks via Joint Disentangled Learning Against Adversarial Attacks}

\author{Canyixing Cui, %
	Tao Wu,~\IEEEmembership{Member,~IEEE}, %
	Xingping Xian, %
	Xiao-Ke Xu,~\IEEEmembership{Member,~IEEE}, %
	Mao Wang, %
	and Weina Niu %
	\thanks{This work was supported in part by the National Natural Science Foundation of China under Grant 62376047; Key Project of Chongqing Natural Science Foundation Innovation and Development Joint Fund under Grant CSTB2023NSCQ-LZX0003; Key Project of Science and Technology Research Program of Chongqing Municipal Education Commission under Grant KJZD-K202300603. \textit{(Corresponding author: Tao Wu)}}%
	\thanks{Canyixing Cui and Mao Wang are with the School of Computer Science and Technology, Chongqing University of Posts and Telecommunications, Chongqing, China. (e-mail: cuicanyixing@163.com; MauStyle@163.com).}
	\thanks{Tao Wu and Xingping Xian are with the School of Cyber Security and Information Law, Chongqing University of Posts and Telecommunications, Chongqing, China. (e-mail: wutaoadeny@gmail.com; xxp0213@gmail.com).}
	\thanks{Xiao-Ke Xu is with the Computational Communication Research Center, Beijing Normal University, Zhuhai, China, and also with the School of Journalism and Communication, Beijing Normal University, Beijing, China. (e-mail: xuxiaoke@foxmail.com).}
	\thanks{Weina Niu is with the School of Computer Science and Engineering, University of Electronic Science and Technology of China, Chengdu, China. (e-mail: vinusniu@uestc.edu.cn).}
	\thanks{The code of GJDNet is available at https://github.com/star4455/GJDNet.}
}

\markboth{Journal of \LaTeX\ Class Files,~Vol.~14, No.~8, August~2021}%
{Cui \MakeLowercase{\textit{et al.}}: GJDNet: Robust Graph Neural Networks via Joint Disentangled Learning Against Adversarial Attacks}


\maketitle

\begin{abstract}
	Graph Neural Networks (GNNs) are vulnerable to adversarial attacks, which inherently invert connectivity patterns by introducing disassortative edges in assortative graphs and assortative edges in disassortative graphs. This structural inversion creates structure–feature mismatches that disrupt neighborhood aggregation across different graph types. However, we find that existing defenses are limited, as they either treat neighborhoods as monolithic under fixed assortativity assumptions or rely on standard softmax classifiers that fail to account for perturbation-induced representation shifts. To further exploit this observation, we adopt a robustness perspective that jointly disentangles node representations and decision spaces, isolating perturbation effects while enforcing well-separated decision regions. Based on this principle, we propose Graph Joint Disentanglement Network (GJDNet), a unified framework for robust node classification across diverse graph assortativity regimes. GJDNet enhances robustness at both representation and decision levels: it employs feature-driven soft structural disentanglement with skewness-aware neighbor filtering to suppress perturbation-induced structure–feature mismatches, and introduces a Spherical Decision Boundary (SDB) to promote intra-class compactness and inter-class separation in the embedding space, thereby stabilizing decision boundaries under perturbations. Theoretical analysis provides insights into the effectiveness of the proposed disentangled representation and decision mechanisms, while extensive experiments demonstrate that GJDNet consistently achieves strong robustness across graphs with different connectivity regimes.
\end{abstract}


\section{Introduction}
\IEEEPARstart{G}{raph} Neural Networks (GNNs) have achieved remarkable success on graph-structured data, including social networks \cite{yu2025aspect}, recommendation systems \cite{zhang2025embedding,lai2023toward}, and biological networks \cite{gligorijevic2021structure,liu2023interpretable}, by iteratively aggregating neighborhood information. Despite their effectiveness, GNNs are highly vulnerable to adversarial attacks, where even small, imperceptible perturbations can disrupt message passing and severely degrade performance, raising critical security and reliability concerns in safety-critical applications \cite{dong2025structure,cao2024multi}.

\begin{figure}[t]
	\centering
	\includegraphics[width=\linewidth]{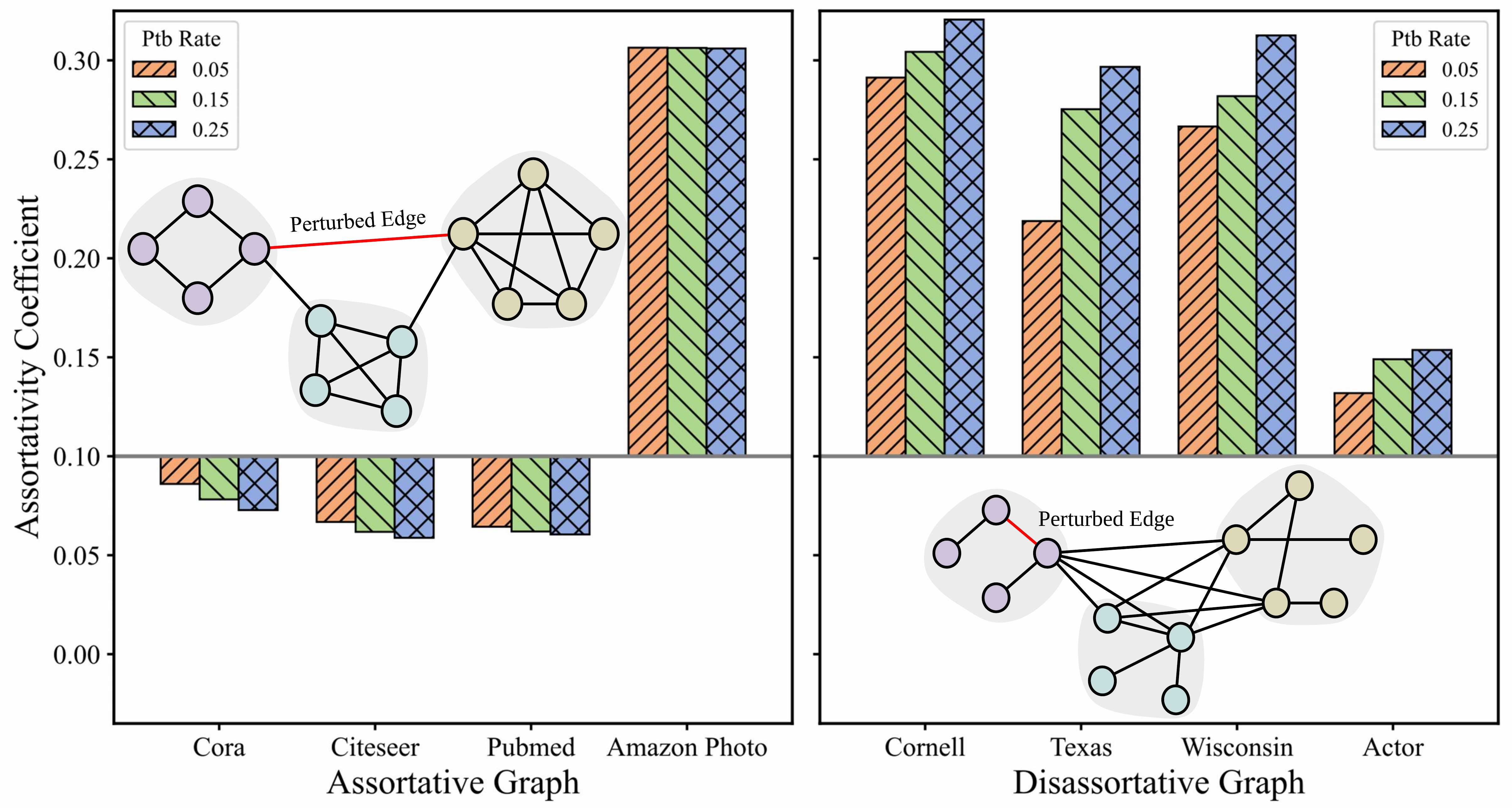}
	\caption{Illustration of assortative (left) and disassortative (right) graphs under structural adversarial attacks. Bars represent feature-based assortativity measured by edge-wise similarity between node features under Min-Max perturbations. Adversarial perturbations invert connectivity patterns, thereby inducing structure–feature mismatches.}	
	\label{fig1_assortativity}
\end{figure}

Extensive research on adversarial attacks against GNNs has proposed a wide range of attack strategies, including gradient-based attacks (e.g., MetaAttack \cite{zugner2019adversarial}, EpoAtk \cite{lin2023exploratory}), reinforcement learning-based attacks (e.g., RL-S2V \cite{dai2018adversarial}, NIPA \cite{sun2020adversarial}, $G^2$-SNIA \cite{chen2024single}), and optimization-based attacks (e.g., TDGIA \cite{zou2021tdgia}, NICKI \cite{sharma2023node}). A common attack mechanism perturbs graph connectivity by introducing edges that invert local connectivity patterns, thereby corrupting message passing \cite{zugner2018adversarial, zugner2019adversarial}. To counteract these attacks, prior defenses have explored several directions, including graph purification methods (e.g., GCN-Jaccard \cite{wu2019adversarial}, GCN-SVD \cite{entezari2020all}), adversarial training approaches (e.g., FLAG \cite{kong2022robust}, GraphAT \cite{feng2019graph}), and robust model designs (e.g., RGCN \cite{zhu2019robust}, SimPGCN \cite{jin2021node}, RUNG \cite{hou2024robust}). These approaches typically modify graph structures, aggregation mechanisms, or regularization schemes to suppress the influence of mismatched edges during message propagation.

Despite these efforts, the fundamental effects of adversarial perturbations on graph connectivity remain underexplored. To investigate this empirically, we analyze adversarial attacks on graphs with different connectivity regimes, including both assortative (homophilic) and disassortative (heterophilic) graphs. The analysis reveals a consistent phenomenon: \textit{structural adversarial attacks tend to invert local connectivity patterns}. Specifically, they introduce disassortative edges in assortative graphs and assortative edges in disassortative graphs (Fig. \ref{fig1_assortativity}), leading to structure–feature mismatches that disrupt neighborhood aggregation. Importantly, this structural inversion occurs across both assortative and disassortative graphs rather than being confined to a specific connectivity regime. This raises a fundamental challenge for robust graph learning: \textit{how can GNNs maintain reliable neighborhood aggregation when adversarial perturbations invert connectivity patterns across different graph regimes?} Meanwhile, adversarial robustness is also closely related to the geometry of decision boundaries. If decision boundaries are overly sensitive or unconstrained, even small perturbations in node representations—particularly for samples near class margins—can lead to misclassification. Moreover, the vulnerability of machine learning models to adversarial attacks often arises from underexplored regions of the feature space, where limited data exposure makes models susceptible to adversarial manipulation. However, existing GNN adversarial defenses largely overlook the distribution of samples in the decision space and fail to impose explicit constraints on class boundaries. This leads to another key question: \textit{how can we design geometry-aware decision mechanisms that stabilize decision boundaries and detect unreliable samples under adversarial perturbations?}

Inspired by disentangled representation learning that aims to separate semantically meaningful latent factors \cite{bengio2013representation, wang2024disentangled}, we advocate a robustness perspective that jointly disentangles node representations and the decision space. By separating latent generative factors within neighborhoods and enforcing structured geometry in the decision space, the model confines the influence of mismatched neighbors to their respective semantic subspaces while stabilizing decision boundaries. Building on this perspective, we propose Graph Joint Disentanglement Network (GJDNet), a unified robust graph learning framework that jointly addresses representation- and decision-level vulnerabilities under structural adversarial attacks. Importantly, GJDNet does not rely on prior assumptions about graph assortativity and can adaptively learn robust representations on graphs with different connectivity regimes. At the representation stage, GJDNet introduces a feature-driven soft structural disentanglement mechanism that projects node features into multiple semantic subspaces and performs subspace-specific neighborhood aggregation, while a skewness-aware filtering strategy suppresses adversarial and long-tail neighbors. At the decision stage, GJDNet employs a spherical decision boundary classifier that models each class as a compact region in the embedding space, enhancing intra-class compactness and inter-class separation. GJDNet mitigates adversarial perturbations during neighborhood aggregation and stabilizes predictions in the decision space.

The main contributions of this work are as follows:
\begin{itemize}
	\item We identify that adversarial attacks inherently invert graph connectivity, creating structure–feature mismatches, and further highlight the role of underexplored regions in the feature space; based on this, we propose a robustness perspective that jointly disentangles node representations and decision spaces to mitigate such effects.
	\item We propose a feature-driven multi-subspace soft structural disentanglement framework with skewness-aware neighbor filtering, which separates latent generative factors within node neighborhoods and effectively suppresses adversarial edges under highly imbalanced similarity distributions.
	\item We introduce a spherical decision boundary formulation that explicitly disentangles the decision space into class-wise regions, promoting intra-class compactness and inter-class separation, thereby enhancing robustness against adversarial perturbations at the decision level.
\end{itemize}

Extensive experiments demonstrate that GJDNet consistently delivers robust performance on graphs with different connectivity regimes, validating the effectiveness of joint representation and decision-space disentanglement.

\section{Related Work}
\subsection{Graph Disentangled Representation Learning}
Graph disentangled representation learning aims to decompose graph representations into multiple latent subspaces that capture distinct semantic or structural factors, thereby improving interpretability and generalization. Existing methods can be categorized into feature-projection-based and structure-pattern-based disentanglement. Early works such as DisenGCN \cite{ma2019disentangled} and IPGDN \cite{liu2020independence} introduced neighborhood routing and statistical independence constraints to disentangle latent factors in graph representations. FactorGCN \cite{yang2020factorizable} factorizes graphs into relation-specific components, while LGD-GCN \cite{guo2022learning} introduces probabilistic modeling for disentanglement. Contrastive learning-based methods such as IDGCL \cite{li2022disentangled} explicitly enforce subspace independence. Beyond node-level representations, HSDN \cite{hu2022hsdn} leverages hypergraph attention and hyperedge-level contrastive learning to capture higher-order semantics. Structure-pattern-based methods focus on disentangling task-relevant and spurious substructures. DisC \cite{fan2022debiasing} and CDGNN \cite{liu2024causal} separate causal and bias subgraphs to mitigate spurious correlations. Despite these advances, most existing disentanglement approaches primarily target interpretability or generalization, leaving open the question of how disentangled representations can be explicitly leveraged to enhance adversarial robustness in graph neural networks.

\subsection{Decision-Space-Based Adversarial Robustness for GNNs}
Studies on adversarial robustness increasingly adopt a decision-space geometry perspective. Early work by Fawzi et al. \cite{fawzi2017robustness} established that robustness is tightly linked to the curvature, smoothness, and proximity of decision boundaries to natural data. Subsequent research further analyzes boundary variability and its impact on generalization \cite{lei2023understanding}. Beyond geometric properties, sensitivity-based and empirical studies demonstrate that spatial transformations, data scale, task complexity, and model architecture jointly shape boundary complexity and adversarial susceptibility \cite{tian2021detecting, sun2019towards}. Some approaches explicitly regularize decision regions through geometric priors, enforcing class-wise or spherical decision regions \cite{zhang2021deep}. Inspired by these insights, graph learning has begun to investigate decision-space robustness. GNNBoundary \cite{wang2024gnnboundary} shows that message passing can induce fragile boundaries, and label scarcity worsens decision separation \cite{ding2024toward}. Other efforts smooth the decision surface via cross-model consistency \cite{deng2025mutual} or extend decision-space ideas to anomaly detection and causal boundary learning \cite{yang2022learning, zhou2023information}. However, few methods explicitly construct geometric decision regions under adversarial attacks. This work addresses this gap by learning class-specific spherical decision regions for robust GNNs.

\subsection{Graph Neural Networks Adversarial Robustness}
Existing studies on adversarial robustness in GNNs can be categorized into graph purification, adversarial training, and robust model design. Graph purification methods aim to sanitize corrupted graph structures by removing or correcting suspicious edges before training. Representative approaches prune unreliable connections using feature similarity or spectral properties (GCN-Jaccard \cite{wu2019adversarial}, GCN-SVD \cite{entezari2020all}). Recent works further refine poisoned graphs through iterative cleaning or generative reconstruction, including FocusedCleaner \cite{zhu2023focusedcleaner} and joint structure-learning methods such as Pro-GNN \cite{jin2020graph}, GraphReshape \cite{wang2024graph}, and REDE \cite{zhang2025adaptive}. Adversarial-training-based defenses enhance robustness by explicitly optimizing GNNs under adversarial perturbations. Typical methods inject perturbations into node features or latent embeddings to enforce adversarial invariance (GraphAT \cite{feng2019graph}; FLAG \cite{kong2022robust}), while others generate adversarial edges or operate in the spectral domain to preserve global smoothness \cite{li2022spectral}. Robust model design methods focus on building inherently stable architectures without explicit purification or adversarial retraining. These include probabilistic modeling of node representations (RGCN \cite{zhu2019robust}), Ricci-curvature-enhanced message passing (Cure-GNN \cite{xiao2022cure}), and similarity-preserving aggregation via auxiliary kNN graphs (SimPGCN \cite{jin2021node}, ERGCN \cite{wu2022ergcn}).

This work bridges graph disentangled representation learning and adversarial robustness in GNNs. Existing disentanglement methods mainly target interpretability or generalization, without explicitly addressing adversarial robustness, while most robust GNN defenses focus on stabilizing message passing and overlook representation disentanglement and decision-space geometry. In contrast, GJDNet jointly disentangles node representations and decision spaces by combining feature-driven soft structural disentanglement with geometry-constrained decision boundaries, providing a unified and principled approach to adversarially robust graph learning.

\section{Preliminaries}
\noindent\textbf{Definition 1 (Node Classification on Graphs).}
Let $\mathcal G=(\mathcal V,\mathcal E)$ denote an undirected graph with $N=|\mathcal V|$ nodes and $E=|\mathcal E|$ edges. The graph is described by an adjacency matrix  $\mathbf{A} \in \mathbb{R}^{N \times N}$ and node features $\mathbf{X} \in \mathbb{R}^{N \times F}$. Each node belongs to one of $C$ classes. In semi-supervised node classification, the node set is partitioned into labeled nodes $\mathcal{V}_l \subset \mathcal{V}$ and unlabeled nodes $\mathcal{V}_u = \mathcal{V} \setminus \mathcal{V}_l$. A GNN $f_\theta$ learns latent node representations from the input graph and produces per-node class logits, from which class probabilities are obtained via the softmax function.

\noindent\textbf{Definition 2 (Robust Graph Learning under Adversarial Attacks).}
Given an attribute graph $\mathcal G$ with ground-truth labels $y$, an adversarial attacker perturbs $\mathcal G$ within a feasible set $\mathcal S(\mathcal G)$, resulting in a perturbed graph $\hat{\mathcal G} \in \mathcal S(\mathcal G)$. Robust graph learning aims to learn parameters $\theta^\ast$ that minimize the worst-case loss over labeled nodes: 
\begin{equation}
	\theta^\ast = \arg\min_{\theta} \max_{\hat{\mathcal G}\in\mathcal S(\mathcal G)} \frac{1}{|\mathcal V_l|} \sum_{i\in\mathcal V_l} \ell \left(f_\theta(\hat{\mathcal G},i), y_i \right),
\end{equation}
where $\ell(\cdot,\cdot)$ denotes the node-level classification loss. This formulation encourages the model to remain stable under adversarial perturbations.

\noindent\textbf{Definition 3 (Graph Disentangled Representation).}
Graph disentangled representation aims to decompose node representations into multiple latent subspaces, each capturing a distinct underlying semantic or generative factor of the graph. Formally, a graph encoder maps the input graph into $K$ subspace-specific representations:
\begin{equation}
	\{\mathbf{H}^{(1)},\ldots,\mathbf{H}^{(K)}\} = \operatorname{Enc}(\mathbf X,\mathbf A), \quad \mathbf{H}^{(k)}\in\mathbb R^{N\times d_k},
\end{equation}
where each $\mathbf{H}^{(k)}$ encodes a semantically coherent component of node features and neighborhood information. The final node representation is obtained by aggregating or concatenating these subspace embeddings, enabling the model to explicitly separate distinct semantic factors within local neighborhoods rather than entangling them into a single embedding.

\noindent\textbf{Definition 4 (Decision Boundary in Embedding Space).}
Let $\mathbf{z}_i \in \mathbb{R}^{d}$ denote the embedding of node $i$, and let $g: \mathbb{R}^{d} \rightarrow \mathbb{R}^{C}$ be the classifier that maps node embeddings to class logits. The decision boundary between classes $c_1$ and $c_2$ is defined as:
\begin{equation}
	\mathcal{B}_{c_1,c_2} = \left\{ \mathbf{z} \in \mathbb{R}^{d} \;\middle|\; g_{c_1}(\mathbf{z}) = g_{c_2}(\mathbf{z}) \right\}.
\end{equation}
This boundary separates regions in the embedding space where class $c_1$ or class $c_2$ is preferred by the classifier. For a labeled node $i$ with ground-truth label $y_i$, the prediction confidence is measured by the logit margin \cite{fawzi2018empirical}:
\begin{equation}
	\mathcal M(i) = g_{y_i}(\mathbf{z}_i) - \max_{c \neq y_i} g_c(\mathbf{z}_i).
\end{equation}
A larger margin indicates that $\mathbf{z}_i$ lies deeper inside the decision region of its true class, whereas a smaller or negative margin implies proximity to, or crossing of, a decision boundary, making the prediction more vulnerable to adversarial perturbations.

\noindent\textbf{Definition 5 (Spherical Decision Region).}
Let $\{\boldsymbol{\mu}_c \in \mathbb{R}^{d}\}_{c=1}^{C}$ denote the class prototypes learned in the embedding space. For each class $c$, a spherical decision region is defined as:
\begin{equation}
	\mathcal{R}_c = \left\{\mathbf{z} \in \mathbb{R}^{d} \;\middle|\; \|\mathbf{z} - \boldsymbol{\mu}_c\|_2^2 \le r_c^2 \right\},
\end{equation}
where $r_c > 0$ is the radius controlling the extent of class $c$. A node $i$ is assigned to class $c$ if its embedding $\mathbf{z}_i$ lies inside $\mathcal{R}_c$ and has the minimum distance to the class center:
\begin{equation}
	\hat y_i = \arg\min_{c} \; \|\mathbf{z}_i - \boldsymbol{\mu}_c\|_2^2.
\end{equation}
Embeddings that fall outside all spherical decision regions are regarded as ambiguous or potentially adversarial, reflecting abnormal positions in the embedding space.

\section{The Proposed GJDNet}

\begin{figure*}[htbp]
	\centering
	\includegraphics[width=\textwidth]{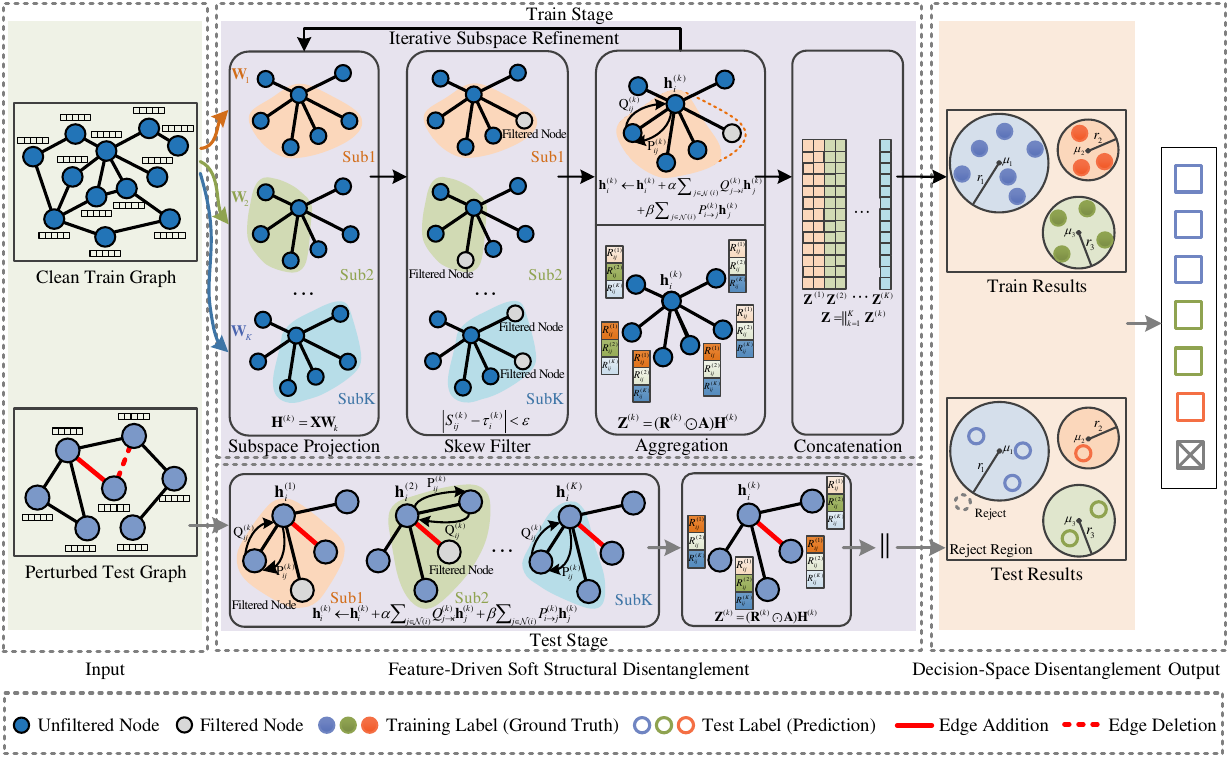}
	\caption{Overview of the proposed GJDNet framework. GJDNet consists of two complementary components: feature-driven soft structural disentanglement and spherical decision-boundary-based decision-space disentanglement. The former performs skewness-aware, feature-driven neighbor filtering to separate heterophilic (mismatched) edges and suppress edges introduced by connectivity pattern inversion, yielding robust node representations. The latter explicitly regularizes class geometry via spherical decision boundaries, thereby enhancing decision robustness against adversarial perturbations.}
	\label{fig2_framework}
\end{figure*}

\subsection{Framework}
The proposed GJDNet enhances robustness against structural adversarial perturbations by jointly stabilizing neighborhood aggregation and decision regions. Specifically, it first applies feature-driven soft structural disentanglement with skewness-aware neighbor filtering to suppress perturbation-induced noise during message passing, and then enforces spherical decision boundary regularization to explicitly shape class-wise geometry in the embedding space. An overview of the framework is illustrated in Fig. \ref{fig2_framework}.

\subsection{Soft Structural Disentanglement for Robust Representation}
The core intuition is that structural adversarial attacks introduce edges that invert local connectivity patterns and create structure–feature mismatches, corrupting neighborhood aggregation and producing unreliable node representations. This vulnerability is aggravated by the monolithic aggregation mechanism of conventional GNNs, where all neighbor messages are indiscriminately mixed within a single shared embedding space. As a result, even a small number of adversarial or mismatched neighbors can influence many feature dimensions simultaneously, allowing local perturbations to propagate across the entire representation space. To address this issue, we decompose neighborhood aggregation into multiple latent subspaces, so that different semantic relations are processed in separate channels rather than being fused into one shared space. Under this design, adversarial neighbors can only affect the subspaces relevant to them instead of contaminating the entire embedding space, thereby preventing local perturbations from propagating globally and improving representation robustness.

\subsubsection{\textit{Feature-driven Soft Structural Disentanglement}}
Following the neighborhood routing mechanism  \cite{ma2019disentangled, zheng2024adversarial}, we assume that the connections between a node and its neighbors are governed by multiple latent generative factors, including those induced by adversarial perturbations. Accordingly, neighbors are softly routed into factor-specific subspaces underlying their connections. The objective of robust learning is to identify, for each node $i$, the subset of neighbors associated with factor $k$, such that aggregation in subspace $k$ involves only neighbors genuinely connected through this factor. This enables accurate modeling of the corresponding semantic component while preventing adversarial neighbors from contaminating the full representation space. The key challenge lies in inferring factor-specific neighbor subsets without explicit supervision. Specifically, we assume $K$ latent generative factors, corresponding to $K$ semantic subspaces projected from the feature space. During subspace-specific aggregation, neighbors irrelevant to factor $k$ are softly down-weighted or pruned. Given the input feature matrix $\mathbf{X} \in \mathbb{R}^{N \times F}$, each node feature $\mathbf{x}_i$ is projected into $K$ latent subspaces via learnable transformations $\mathbf{W}_k \in \mathbb{R}^{F \times d_k}$. The subspace embeddings are $\ell_2$-normalized to eliminate scale variations:
\begin{equation}
	\mathbf{h}_i^{(k)} = \frac{\mathbf{x}_i \mathbf{W}_k}{\left\| \mathbf{x}_i \mathbf{W}_k \right\|_2} \in \mathbb{R}^{d_k}.
	\label{eq:subspaceproject}
\end{equation}  
For each subspace $k$, the interaction between node $i$ and its neighbor $j \in \mathcal{N}(i)$ is quantified as:
\begin{equation}
	S_{ij}^{(k)} =	\cos\bigl(\mathbf{h}_i^{(k)}, \mathbf{h}_j^{(k)}\bigr) = \frac{\bigl\langle \mathbf{h}_i^{(k)}, \mathbf{h}_j^{(k)} \bigr\rangle} {\|\mathbf{h}_i^{(k)}\| \cdot \|\mathbf{h}_j^{(k)}\|}.
	\label{eq:cosinesim}
\end{equation}

\subsubsection{\textit{Skewness-Aware Adversarial Neighbor Filtering}}
\begin{figure}[t]
	\centering
	\subfloat[Left Skewness]{%
		\includegraphics[width=0.48\linewidth]{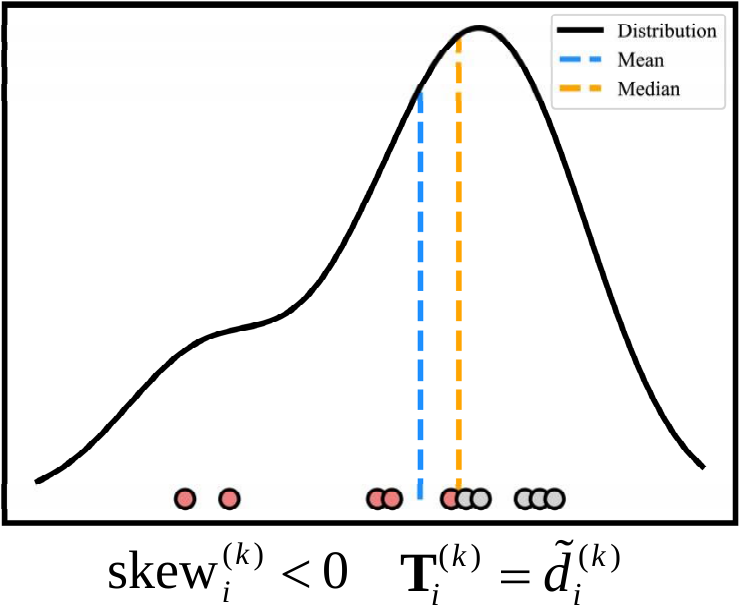}%
		\label{fig3_skewfilter_a}
	}
	\hfill
	\subfloat[Right Skewness]{%
		\includegraphics[width=0.48\linewidth]{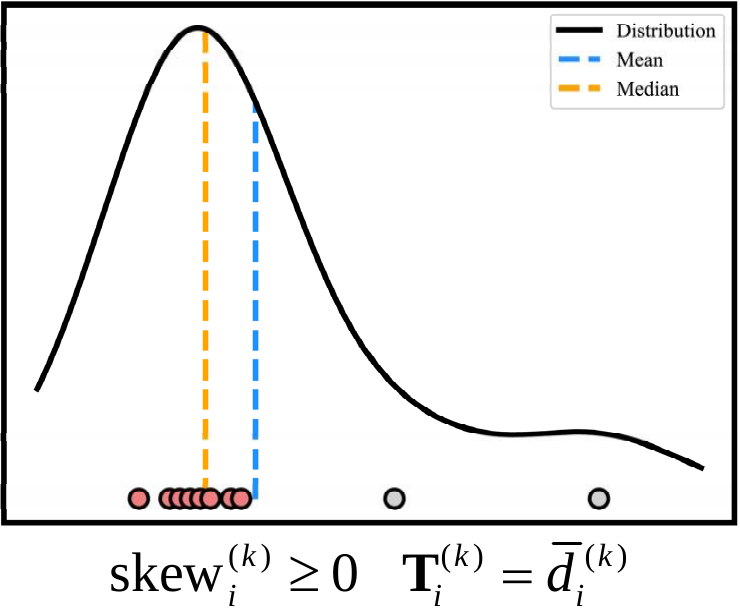}%
		\label{fig3_skewfilter_b}
	}	
	\caption{Skewness-aware neighbor filtering based on local similarity distributions. (a) For left-skewed local similarity distributions ($\mathrm{skew}_i^{(k)} < 0$), the median $\tilde{d}_i^{(k)}$ is selected as the adaptive threshold. (b) For right-skewed distributions ($\mathrm{skew}_i^{(k)} \ge 0$), the mean $\bar{d}_i^{(k)}$ is used instead.}
	\label{fig3_skewfilter}
\end{figure}

Under structural adversarial perturbations, the neighborhood similarity distribution often becomes asymmetric due to a small number of extreme perturbed neighbors, manifesting as left-skewed distributions caused by low-similarity outliers or right-skewed distributions dominated by abnormally high similarities (Fig. \ref{fig3_skewfilter}). Despite their small number, these extreme neighbors can exert disproportionate influence during aggregation, significantly degrading node representations. Since attention weights are normalized within each neighborhood, extreme similarity values can dominate the aggregation process, allowing a small number of perturbed edges to disproportionately control representation updates. To address this issue, we characterize such outliers via the skewness of the similarity distribution and adopt a robustness-oriented measure defined as the difference between the mean and the median:
\begin{equation}
	\mathrm{skew}_i^{(k)} = \bar d_i^{(k)} - \tilde d_i^{(k)},
	\label{eq:skew}
\end{equation}
where $\bar d_i^{(k)}$ and $\tilde d_i^{(k)}$ denote the mean and median of the similarity set $\{{S}_{ij}^{(k)} \mid j \in \mathcal{N}(i)\}$, respectively. Negative values indicate left-skewed similarity distributions caused by low-similarity outliers, while non-negative values correspond to right-skewed distributions dominated by highly similar neighbors. Based on this observation, we define an adaptive threshold for each subspace as:
\begin{equation}
	{\tau}_i^{(k)} = 
	\begin{cases} 
		\bar d_i^{(k)}, & \mathrm{skew}_i^{(k)} \ge 0, \\ 
		\tilde d_i^{(k)}, & \mathrm{skew}_i^{(k)} < 0.
	\end{cases}
	\label{eq:Tik}
\end{equation}

This design aligns with the two skewness regimes illustrated in Fig. \ref{fig3_skewfilter} and enables robust thresholding under asymmetric local similarity distributions. Neighbors whose similarity scores deviate from the adaptive threshold by more than a tolerance margin $\varepsilon$ are treated as unreliable outliers and filtered out:
\begin{equation}
	{M}_{ij}^{(k)} = 
	\begin{cases}
		1, & \left| {S}_{ij}^{(k)} - {\tau}_i^{(k)} \right| < \varepsilon, \\
		0, & \text{otherwise}.
	\end{cases}
	\label{eq:skewmask}
\end{equation}
To enforce hard exclusion in the intra-subspace attention refinement, we define masked logits as:
\begin{equation}
	\tilde{{S}}_{ij}^{(k)}=
	\begin{cases}
		{S}_{ij}^{(k)}, & {M}_{ij}^{(k)}=1,\\
		-\infty, & {M}_{ij}^{(k)}=0.
	\end{cases}
	\label{eq:masked_logits}
\end{equation}
Consequently, neighbors connected by masked edges are excluded from aggregation in that subspace.

\subsubsection{\textit{Cross-Subspace Routing for Robust Aggregation}}
Although semantic subspaces enable factor-specific aggregation, latent generative factors are not strictly independent in practice. A neighbor may influence a target node through multiple semantic factors rather than a single subspace. To capture this dependency, we introduce a soft cross-subspace routing mechanism that allows each edge to adaptively distribute its contribution across multiple subspaces. Specifically, within each subspace, we introduce dual directional attention mechanisms to capture fine-grained semantic interactions. The incoming attention $Q_{j \rightarrow i}^{(k)}$ measures the influence of neighbor $j$ on center node $i$, normalized over the neighborhood of node $i$, while the outgoing attention $P_{i \rightarrow j}^{(k)}$ characterizes the same edge influence normalized over the neighborhood of node $j$:
\begin{equation}
	Q_{j \rightarrow i}^{(k)} = \frac{\exp(\tilde{{S}}_{ij}^{(k)})}	{\sum\limits_{j' \in {\mathcal N}(i)} \exp(\tilde{{S}}_{ij'}^{(k)})},	
	\quad
	P_{i \rightarrow j}^{(k)} = \frac{\exp(\tilde{{S}}_{ij}^{(k)})}	{\sum\limits_{i' \in {\mathcal N}(j)} \exp(\tilde{{S}}_{i'j}^{(k)})}.
	\label{eq:masked_attention}
\end{equation}
To avoid unstable softmax normalization when masking yields empty or near-empty neighborhoods on low-degree nodes, skew filtering is disabled when the node degree is below ${min\_neighbors}$. Based on the incoming and outgoing attention weights $Q_{j \rightarrow i}^{(k)}$ and $P_{i \rightarrow j}^{(k)}$, we implement a dynamic subspace assignment mechanism via iterative refinement. Starting from the initial subspace representations $\mathbf{h}_i^{(k)}$, the aggregated representation of node $i$ in subspace $k$ is updated as:
\begin{equation}
	\mathbf{h}_i^{(k)} \leftarrow \mathbf{h}_i^{(k)} + \alpha \sum_{j \in \mathcal{N}(i)} Q_{j \rightarrow i}^{(k)} \mathbf{h}_j^{(k)} + \beta \sum_{j \in \mathcal{N}(i)} P_{i \rightarrow j}^{(k)} \mathbf{h}_j^{(k)},
	\label{eq:node_agg}
\end{equation}
where $\alpha$ and $\beta$ control the strengths of incoming and outgoing directional aggregation, respectively. Through iterative updates, neighbors are softly routed into factor-specific subspaces according to their directional semantic relevance, yielding robust subspace representations that jointly preserve factor disentanglement and capture cross-factor interactions.

Upon node-level refinement, we conduct subspace-wise structure-aware aggregation, in which neighbor contributions are adaptively modulated by learned cross-subspace routing weights. This design enables each semantic subspace to selectively emphasize factor-relevant neighborhood information while explicitly modeling dependencies among latent factors. Specifically, for the $k$-th semantic subspace, node representations are aggregated as:
\begin{equation}
	\mathbf{Z}^{(k)} = \left(\mathbf{R}^{(k)} \odot \mathbf{A}\right)\mathbf{H}^{(k)},
	\label{eq:subspace_agg}
\end{equation}
where $\mathbf{R}^{(k)}$ denotes the routing weight matrix encoding the subspace-specific contribution of each edge, with:
\begin{equation}
	{R}_{ij}^{(k)} = \frac{\exp\!\left({S}_{ij}^{(k)}\right)}{\sum_{l=1}^K \exp\!\left({S}_{ij}^{(l)}\right)},
	\label{eq:routing_softmax}
\end{equation}
where routing weights are computed from the unmasked similarities ${S}_{ij}^{(k)}$ to ensure stable cross-subspace normalization, while masking is applied only to intra-subspace attention to hard-exclude outliers during neighborhood aggregation.

The final node representation is obtained by concatenating the outputs from all $K$ semantic subspaces:
\begin{equation}
	\mathbf{Z} = \left[ \mathbf{Z}^{(1)} \| \mathbf{Z}^{(2)} \| \cdots \| \mathbf{Z}^{(K)} \right],
	\quad d = \sum_{k=1}^{K} d_k,
	\label{eq:Z_concat}
\end{equation}
thereby jointly encoding multiple latent semantic factors. The proposed aggregation captures complementary structural and semantic patterns across subspaces while enabling information sharing via soft routing, resulting in robust node representations. The overall procedure is summarized in Algorithm \ref{alg:disentangle}.

\begin{algorithm}[htbp]
	\caption{Feature-Driven Soft Structural Disentanglement}
	\label{alg:disentangle}
	\textbf{Input:} Graph $\mathcal{G}=(\mathcal{V},\mathcal{E})$; features $\mathbf{X}$; projections $\{\mathbf{W}_k\}_{k=1}^K$; adjacency $\mathbf{A}$; margin $\varepsilon$; coefficients $\alpha,\beta$; iterations $T$ \\
	\textbf{Output:} Node representation $\mathbf{Z}$
	\begin{algorithmic}[1]
		\FOR{$k = 1$ to $K$}
		\STATE Project features and initialize $\mathbf{H}^{(k)}$ (Eq. \eqref{eq:subspaceproject})
		\FOR{iteration $t = 1$ to $T$}
		\STATE Compute similarities $\mathbf{S}^{(k)}$ (Eq. \eqref{eq:cosinesim})
		\STATE Obtain skewness-aware mask $\mathbf{M}^{(k)}$ (Eqs. \eqref{eq:skew}--\eqref{eq:skewmask})
		\STATE Mask similarities $\tilde{\mathbf{S}}^{(k)}$ (Eq. \eqref{eq:masked_logits})
		\STATE Compute directional attentions $\mathbf{Q}^{(k)}, \mathbf{P}^{(k)}$ (Eq. \eqref{eq:masked_attention})
		\STATE Update subspace embeddings $\mathbf{H}^{(k)}$ (Eq. \eqref{eq:node_agg})
		\ENDFOR
		\ENDFOR
		\STATE Compute routing weights $\{\mathbf{R}^{(k)}\}_{k=1}^K$ (Eq. \eqref{eq:routing_softmax})
		\FOR{$k = 1$ to $K$}
		\STATE Aggregate $\mathbf{Z}^{(k)} = (\mathbf{R}^{(k)} \odot \mathbf{A}) \mathbf{H}^{(k)}$
		\ENDFOR
		\STATE $\mathbf{Z} = [\mathbf{Z}^{(1)} \| \cdots \| \mathbf{Z}^{(K)}]$
		\RETURN $\mathbf{Z}$
	\end{algorithmic}
\end{algorithm}

\subsection{Spherical Decision Boundary (SDB)}
\begin{figure}[htbp]
	\centering
	\subfloat[$\mathcal{L}_{\mathrm{CE}}$]{
		\includegraphics[width=0.42\linewidth]{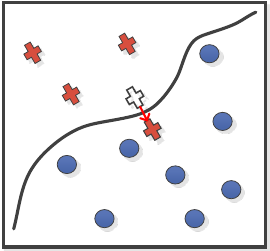}
		\label{fig4_boundary_a}}
	\hfill
	\subfloat[${{\mathcal L}_{CE}} + {{\mathcal L}_{\mathrm{SDB}}}$]{
		\includegraphics[width=0.42\linewidth]{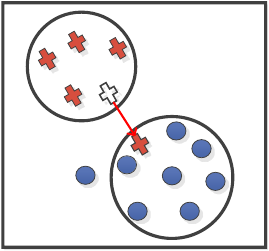}
		\label{fig4_boundary_b}}	
	\caption{Illustration of the SDB mechanism. (a) Under cross-entropy loss $\mathcal{L}_{\mathrm{CE}}$, unconstrained decision boundaries allow adversarial perturbations to easily cross class boundaries (red arrow). (b) The spherical regularization $\mathcal{L}_{\mathrm{SDB}}$ confines samples within compact class regions, enlarging margins and improving robustness. The cross denotes a perturbed sample.}
	\label{fig4_boundary}
\end{figure}

Adversarial attacks can undermine GNNs at the decision stage by distorting the global geometry of the embedding space, leading to unstable decision boundaries even with robust representations. In particular, adversarial perturbations may push node embeddings toward low-density regions or across class boundaries, thereby amplifying classification uncertainty. To explicitly address decision-level vulnerability, we propose a Spherical Decision Boundary (SDB) learning mechanism that explicitly regularizes embedding geometry by enforcing class-wise compactness and inter-class separation. Such geometric regularization yields smoother and more stable decision boundaries, as illustrated in Fig. \ref{fig4_boundary}. By organizing node embeddings into class-specific spherical regions, SDB induces smooth and stable decision boundaries and forces adversarial perturbations to cross explicit class-wise boundaries, increasing attack difficulty. Moreover, the induced geometry naturally enables a rejection mechanism, whereby embeddings inconsistent with all class-specific regions are identified as low-confidence or adversarial samples and rejected. In this way, SDB complements robust representation learning with explicit geometry-aware decision modeling.

Given the node representation matrix $\mathbf{Z}\in\mathbb{R}^{N\times d}$ produced by the soft disentanglement module, where $\mathbf{z}_i$ denotes the embedding of node $i$, we define the labeled embedding set of class $c$ as $\mathcal{Z}_c = \{\mathbf{z}_i \mid i \in \mathcal{V}_l,\, y_i = c\}$, where $\mathcal{V}_l$ is the set of labeled nodes. The centroid of class $c$ is computed as the mean embedding of its labeled samples, $\boldsymbol{\mu}_c = \frac{1}{|\mathcal{Z}_c|} \sum_{\mathbf{z}_i \in \mathcal{Z}_c} \mathbf{z}_i$. To preserve predictive capability, we first adopt the cross-entropy loss to supervise node classification: $\mathcal{L}_{\mathrm{CE}} = - \sum_{i \in \mathcal{T}} \log p_{i,y_i}$, where $p_{i,y_i}$ denotes the predicted probability that node $i$ is assigned to its ground-truth class $y_i$. Beyond conventional supervision, the proposed SDB is realized through two complementary geometric objectives that enforce class-wise decision regions. Specifically, the intra-class compactness loss penalizes large deviations of node embeddings from their corresponding class centroids, encouraging embeddings of the same class to concentrate within compact spherical regions:
\begin{equation}
	\mathcal{L}_{\mathrm{intra}} = \frac{1}{|\mathcal{C}|} \sum_{c \in \mathcal{C}} \frac{1}{|\mathcal{Z}_c|} \sum_{\mathbf{z}_i \in \mathcal{Z}_c} \left\| \mathbf{z}_i - \boldsymbol{\mu}_c \right\|_2^2.
\end{equation}
This term reduces intra-class variance and limits the influence of adversarially perturbed outliers within each class. To further enlarge the margin between different classes, we introduce an inter-class separation loss that encourages class centroids to be well separated in the embedding space:
\begin{equation}
	\mathcal{L}_{\mathrm{inter}} = \frac{2}{|\mathcal{C}|(|\mathcal{C}| - 1)} \sum_{c_1 < c_2} \left\| \boldsymbol{\mu}_{c_1} - \boldsymbol{\mu}_{c_2} \right\|_2^2,
\end{equation}
where $\mathcal{C}$ denotes the set of classes. By explicitly maximizing pairwise distances between class centroids, this objective enlarges the effective inter-class margins between spherical decision regions. The intra-class compactness and inter-class separation objectives are jointly integrated into a unified spherical decision boundary regularization term:
\begin{equation}
	\mathcal{L}_{\mathrm{SDB}} = \frac{\mathcal{L}_{\mathrm{intra}}} {\mathcal{L}_{\mathrm{inter}} + \delta},
\end{equation}
where $\delta > 0$ is a small constant introduced for numerical stability. This ratio-based formulation provides a scale-invariant balance between intra-class compactness and inter-class separation, effectively preventing trivial solutions caused by uniformly shrinking or expanding the embedding space.

Finally, the training objective jointly optimizes the classification loss and spherical decision boundary regularization:
\begin{equation}
	\mathcal{L} = \mathcal{L}_{\mathrm{CE}} + \lambda \, \mathcal{L}_{\mathrm{SDB}},
\end{equation}
where $\lambda$ controls the strength of the geometry-aware regularization. By explicitly shaping the embedding space with class-specific spherical constraints, the proposed SDB mechanism introduces a geometry-aware inductive bias that stabilizes decision boundaries under adversarial perturbations. 

\subsection{SDB-Guided Decision Mechanism}
To complement the SDB’s geometric regularization, we employ a sphere-based inference mechanism that predicts classes based on explicit spherical decision regions. For each class $c \in \mathcal{C}$, the region is defined as a closed ball centered at $\boldsymbol{\mu}_c$, with radius $r_c = \max_{\mathbf{z}_i \in \mathcal{Z}_c}\|\mathbf{z}_i - \boldsymbol{\mu}_c\|_2$ for simplicity and interpretability. For a test embedding $\mathbf{z} \in \mathbb{R}^d$, the set of candidate classes whose spherical regions contain it is:
\begin{equation}
	\mathcal{C}_z = \left\{ c \in \mathcal{C} \;\middle|\; \left\| \mathbf{z} - \boldsymbol{\mu}_c \right\|_2 \le r_c  \right\}.
\end{equation}
The predicted label $\hat{y}$ is then determined by:
\begin{equation}
	\hat{y} =
	\begin{cases}
		\text{reject}, & \mathcal{C}_z = \emptyset, \\[4pt]
		c, & \mathcal{C}_z = \{c\}, \\[4pt]
		\arg\min\limits_{c \in \mathcal{C}_z} \left\| \mathbf{z} - \boldsymbol{\mu}_c \right\|_2, & |\mathcal{C}_z| > 1 .
	\end{cases}
\end{equation}
Accordingly, the rejection region is explicitly defined as:
\begin{equation}
	\mathcal{R}_{\mathrm{reject}} = \mathbb{R}^d \setminus \bigcup_{c \in \mathcal{C}} B(\boldsymbol{\mu}_c, r_c),
\end{equation}
where $B(\boldsymbol{\mu}_c, r_c) = \left\{ \mathbf{z} \;\middle|\; \left\| \mathbf{z} - \boldsymbol{\mu}_c \right\|_2 \le r_c \right\}$ denotes the closed ball associated with class $c$.

\subsection{Property Analysis}
\subsubsection{Property 1. Stability of Soft Disentangled Aggregation}
\label{sec:PA_property1}
Let $\mathbf{A}$ and $\mathbf{A}'$ denote the normalized adjacency matrices
before and after a structural perturbation, respectively, and define $\Delta\mathbf{A}\triangleq \mathbf{A}'-\mathbf{A}$. According to Eq. \eqref{eq:subspace_agg}, for factor $k$, the subspace-wise aggregation is:
\begin{equation}
	\mathbf{Z}^{(k)} = \big(\mathbf{A}\odot \mathbf{R}^{(k)}\big)\mathbf{H}^{(k)},
	\quad
	\mathbf{Z}'^{(k)} = \big(\mathbf{A}'\odot \mathbf{R}'^{(k)}\big)\mathbf{H}'^{(k)},
	\label{eq:PA_Zk_def}
\end{equation}
where $\mathbf{H}^{(k)}$ and $\mathbf{H}'^{(k)}$ denote the attention-refined subspace representations obtained via Eqs. \eqref{eq:masked_attention} and \eqref{eq:node_agg}, and $\mathbf{R}^{(k)}$ and $\mathbf{R}'^{(k)}$ are the corresponding routing weight matrices computed by Eq. \eqref{eq:routing_softmax}. The aggregated node representations before and after the structural perturbation are given by $\mathbf{Z} = [\mathbf{Z}^{(1)} \| \cdots \| \mathbf{Z}^{(K)}]$ and $\mathbf{Z}' = [\mathbf{Z}'^{(1)} \| \cdots \| \mathbf{Z}'^{(K)}]$.

\paragraph{Non-Amplifying Routing of Soft Disentanglement}
For edge $(i,j)$, the routing weights satisfy a simplex constraint, i.e., $R_{ij}^{(k)} \in [0,1]$ and $\sum_{k=1}^{K} R_{ij}^{(k)} = 1$, which implies that routing does not amplify perturbations across subspaces. Specifically, for any matrix $\mathbf{X}$, since $\sum_{k} (R_{ij}^{(k)})^2 \le (\sum_{k} R_{ij}^{(k)})^2 = 1$,
\begin{equation}
	\sum_{k=1}^{K} \|\mathbf{X} \odot \mathbf{R}^{(k)}\|_F^2 = \sum_{i,j} X_{ij}^2 \sum_{k=1}^{K} (R_{ij}^{(k)})^2 \le \|\mathbf{X}\|_F^2,
	\label{eq:PA_nonamp_general}
\end{equation}
Let $\mathbf{X} = \Delta \mathbf{A}$, $ \sum_{k=1}^{K} \|\Delta \mathbf{A} \odot \mathbf{R}^{(k)}\|_F^2 \le \|\Delta \mathbf{A}\|_F^2$, which shows that adversarial perturbations cannot be amplified by the routing mechanism. This property suggests that the routing mechanism does not amplify structural perturbations across subspaces, promoting stable disentangled aggregation.

\paragraph{Stability of Aggregation Mechanism under Structural perturbations}
For each factor $k$, the representation shift induced by an adversarial perturbation can be decomposed as:
\begin{equation}
	\begin{aligned}
		\mathbf{Z}'^{(k)}-\mathbf{Z}^{(k)}= \;& \big((\Delta\mathbf{A})\odot \mathbf{R}^{(k)}\big)\mathbf{H}^{(k)} + \big(\mathbf{A}'\odot (\mathbf{R}'^{(k)} \\ -\mathbf{R}^{(k)})\big)\mathbf{H}^{(k)} &+ \big(\mathbf{A}'\odot  \mathbf{R}'^{(k)}\big)\big(\mathbf{H}'^{(k)}-\mathbf{H}^{(k)}\big),
	\end{aligned}
	\label{eq:PA_three_term_decomp}
\end{equation}
which explicitly separates the effects of structural perturbation, routing drift, and embedding drift. Applying the triangle inequality and $\|XY\|_F\le \|X\|_2\|Y\|_F$ yields:
\begin{equation}
	\begin{aligned}
		\|\mathbf{Z}'^{(k)}-\mathbf{Z}^{(k)}\|_F \le\;&	\underbrace{ \|(\Delta\mathbf{A})\odot \mathbf{R}^{(k)}\|_2\, \|\mathbf{H}^{(k)}\|_F }_{\text{structural perturbation}} \\
		&+
		\underbrace{ \|\mathbf{A}'\odot (\mathbf{R}'^{(k)}-\mathbf{R}^{(k)})\|_2\, \|\mathbf{H}^{(k)}\|_F }_{\text{routing drift}} \\
		&+
		\underbrace{ \|\mathbf{A}'\odot \mathbf{R}'^{(k)}\|_2\, \|\mathbf{H}'^{(k)}-\mathbf{H}^{(k)}\|_F }_{\text{embedding drift}}
	\end{aligned}.
	\label{eq:PA_three_term_singlek}
\end{equation}
Moreover, the final representation is obtained by concatenation across subspaces:
\begin{equation}
	\|\mathbf{Z}'-\mathbf{Z}\|_F^2 = \sum_{k=1}^{K}\|\mathbf{Z}'^{(k)}-\mathbf{Z}^{(k)}\|_F^{2}.
	\label{eq:PA_concat_identity}
\end{equation}
This additive bound shows that the total representation shift can be explicitly controlled across subspaces. As a consequence, the proposed soft structural disentanglement guarantees that adversarial perturbations remain bounded throughout the aggregation process, providing a theoretical stability condition for robust message passing under structural attacks.

\subsubsection{Property 2. Universality of Spherical Decision Boundaries}
While SDB employs isotropic spherical regions for simplicity and optimization stability, its formulation naturally extends to anisotropic ellipsoidal regions under a Mahalanobis metric, since assuming perfectly spherical class regions is often unrealistic in real-world settings. For a node embedding $\mathbf{z}\in\mathbb{R}^d$, consider the generalized class-wise distance:
\begin{equation}
	D_c(\mathbf{z}) \triangleq (\mathbf{z}-\boldsymbol{\mu}_c)^\top \mathbf{M}_c (\mathbf{z}-\boldsymbol{\mu}_c), \quad \mathbf{M}_c \succ \mathbf{0},
	\label{eq:PA_mahalanobis_main}
\end{equation}
where $\mathbf{M}_c \succ \mathbf{0}$ is a symmetric positive definite matrix that induces a valid Mahalanobis metric. The inequality $D_c(\mathbf{z}) \le \rho_c$ defines an ellipsoidal decision region:
\begin{equation}
	\widetilde{\mathcal{R}}_c \triangleq \left\{\mathbf{z}\in\mathbb{R}^d \ \middle|\ D_c(\mathbf{z}) \le \rho_c \right\}, \quad \rho_c>0,
	\label{eq:PA_ellipsoid_main}
\end{equation}
where $\rho_c$ controls the scale of the class decision region.

The spherical formulation used in SDB is recovered as the isotropic case $\mathbf{M}_c=\mathbf{I}$, where $D_c(\mathbf{z})=\|\mathbf{z}-\boldsymbol{\mu}_c\|_2^2$. Setting $\rho_c=r_c^2$ yields the Euclidean ball $\{\mathbf{z}\mid \|\mathbf{z}-\boldsymbol{\mu}_c\|_2 \le r_c\}$ used by the spherical decision regions. The above derivation shows that spherical decision regions can be viewed as a special case of a more general ellipsoidal formulation. Nevertheless, learning a full ellipsoidal metric requires estimating $\mathcal{O}(d^2)$ parameters per class and enforcing positive definiteness, which substantially increases computational and statistical complexity and may lead to optimization instability and sensitivity to anisotropic directions in high-dimensional settings. Therefore, we adopt the spherical formulation in this work as a robust and computationally efficient approximation.

\section{Experiments}
We conduct comprehensive experiments to systematically evaluate the robustness, effectiveness, and efficiency of GJDNet by addressing the following research questions:
\begin{itemize}
	\item \textbf{RQ1}
	How does GJDNet perform against state-of-the-art baselines under structural perturbations across graphs with different connectivity regimes?
	\item \textbf{RQ2}
	How do the representations and decision boundaries of GJDNet behave under adversarial perturbations?
	\item \textbf{RQ3}
	How much does each proposed component contribute to the robustness of GJDNet?
	\item \textbf{RQ4}
	How do key hyperparameters influence the performance of GJDNet?
	\item \textbf{RQ5}
	What is the computational complexity and empirical running-time cost of GJDNet?
\end{itemize}

\subsection{Experimental Settings}
\textbf{Datasets.}
We evaluate GJDNet on eight benchmark datasets: Cora, Citeseer, Pubmed, Amazon Photo, Cornell, Texas, Wisconsin, and Actor. The first four are assortative citation/co-purchase graphs, while the latter four are disassortative graphs. Dataset statistics are reported in Table \ref{datasets}.

\begin{table}[!h]
	\caption{Node classification datasets}
	\centering
	\renewcommand{\arraystretch}{1.15}
	\resizebox{1.0 \columnwidth}{!}{
		\begin{tabular}{cccccc}
			\hline
			\textbf{Dataset} & \textbf{Nodes} & \textbf{Edges} & \textbf{Features} & \textbf{Classes} & 
			\textbf{Assortativity}  \\
			\hline
			\textbf{Cora}     & 2,485  & 5,069 & 1,433 & 7 & 0.83  \\
			\rowcolor{gray!15}
			\textbf{Citeseer} & 2,110  & 3,668 & 3,703 & 6 & 0.71  \\
			\textbf{Pubmed}   & 19,717 & 44,338 & 500 & 3 & 0.79  \\
			\rowcolor{gray!15}
			\textbf{Amazon Photo}   & 7,650 & 238,162 & 745 & 8 & 0.84  \\
			\hline
			\textbf{Cornell}   & 183 & 298 & 1,703 & 5 & 0.11 \\
			\rowcolor{gray!15}
			\textbf{Texas}   & 183 & 325 & 1,703 & 5 & 0.06 \\
			\textbf{Wisconsin}   & 251 & 515 & 1,703 & 5 & 0.16  \\
			\rowcolor{gray!15}
			\textbf{Actor}   & 7,600 & 30,019 & 932 & 5 & 0.24  \\
			\hline
		\end{tabular}
	}
	\label{datasets}
\end{table}

\textbf{Baselines.}
We compare GJDNet with standard message-passing GNN backbones and representative robustness methods: GCN \cite{kipf2016semi}, GAT \cite{velivckovic2017graph}, GCN-Jaccard \cite{wu2019adversarial}, RGCN \cite{zhu2019robust}, GCN-SVD \cite{entezari2020all}, SimPGCN \cite{jin2021node}, ERGCN \cite{wu2022ergcn}, ADGCN \cite{zheng2024adversarial}, and GraphReshape \cite{wang2024graph}.

\textbf{Attacks.}
To evaluate robustness under structural perturbations, we consider three attack methods:
(i) Min-Max \cite{xu2019topology}, an untargeted global topology attack that maximizes classification loss under structural constraints;
(ii) Nettack \cite{zugner2018adversarial}, a targeted node-level attack manipulating local graph structure, where feature perturbations are disabled;
(iii) Random Attack \cite{malik2017robustness}, randomly adding or removing edges to simulate noise.

\textbf{Implementation Details.}
All methods use a 2-layer architecture and are trained for 200 epochs with ReLU, dropout 0.5, learning rate 0.01, and weight decay $5\times10^{-4}$. Results are averaged over 10 runs. Experiments are implemented in PyTorch 1.10.1 and run on an RTX 4090 (CUDA 11.3). For GJDNet, we set $K=3$.

\begin{table*}[!t]
	\centering
	\caption{Accuracy of node classification methods under Min-Max attacks on assortative and disassortative graphs.}
	\label{tab2_minmax}
	{\fontsize{7pt}{6pt}\selectfont
		\renewcommand{\arraystretch}{1.4}
		\subfloat[Accuracy under Min-Max attack on assortative graphs]{\resizebox{\textwidth}{!}
			{\begin{tabular}{c|c|cccccccccc}
					\hline
					\textbf{Dataset} & \textbf{Ptb Rate} & \textbf{GCN} & \textbf{GAT} & \textbf{GCN-Jaccard} & \textbf{RGCN} & \textbf{GCN-SVD} & \textbf{SimPGCN} & \textbf{ERGCN} & \textbf{ADGCN} & \textbf{GraphReshape} & \textbf{GJDNet} \\
					\hline
					\multirow{6}{*}{\textbf{Cora}}  & 0.00 & 0.8325 & 0.8407 & 0.8370 & 0.8422 & 0.8271 & 0.8287 & 0.8452 & 0.8096 & 0.8410 & \textbf{0.8564}\\
					
					& \cellcolor{gray!15}0.05 & \cellcolor{gray!15}0.8252 & \cellcolor{gray!15}0.8347 & \cellcolor{gray!15}0.8298 & \cellcolor{gray!15}0.8351 & \cellcolor{gray!15}0.8177 & \cellcolor{gray!15}0.8155 & \cellcolor{gray!15}0.8357 & \cellcolor{gray!15}0.7987 & \cellcolor{gray!15}0.8332 & \cellcolor{gray!15}\textbf{0.8547}\\
					& 0.10 & 0.7995 & 0.8102 & 0.8035 & 0.8106 & 0.7946 & 0.7981 & 0.8119 & 0.7859 & 0.8067 & \textbf{0.8329}\\
					& \cellcolor{gray!15}0.15 & \cellcolor{gray!15}0.7803 & \cellcolor{gray!15}0.7886 & \cellcolor{gray!15}0.7851 & \cellcolor{gray!15}0.7893 & \cellcolor{gray!15}0.7738 & \cellcolor{gray!15}0.7828 & \cellcolor{gray!15}0.7902 & \cellcolor{gray!15}0.7730 & \cellcolor{gray!15}0.7871 & \cellcolor{gray!15}\textbf{0.8128}\\
					& 0.20 & 0.7616 & 0.7695 & 0.7651 & 0.7697 & 0.7584 & 0.7670 & 0.7769 & 0.7564 & 0.7674 & \textbf{0.7947}\\
					& \cellcolor{gray!15}0.25 & \cellcolor{gray!15}0.7388 & \cellcolor{gray!15}0.7440 & \cellcolor{gray!15}0.7422 & \cellcolor{gray!15}0.7450 & \cellcolor{gray!15}0.7410 & \cellcolor{gray!15}0.7520 & \cellcolor{gray!15}0.7473 & \cellcolor{gray!15}0.7432 & \cellcolor{gray!15}0.7442 & \cellcolor{gray!15}\textbf{0.7770}\\
					\hline
					\multirow{6}{*}{\textbf{Citeseer}} & 0.00 & 0.7228 & 0.7294 & 0.7267 & 0.7270 & 0.7392 & 0.7397 & 0.7402 & 0.7286 & 0.7239 & \textbf{0.7651} \\
					& \cellcolor{gray!15}0.05 & \cellcolor{gray!15}0.7202 & \cellcolor{gray!15}0.7250 & \cellcolor{gray!15}0.7238 & \cellcolor{gray!15}0.7221 & \cellcolor{gray!15}0.7380 & \cellcolor{gray!15}0.7360 & \cellcolor{gray!15}0.7374 & \cellcolor{gray!15}0.7243 & \cellcolor{gray!15}0.7206 & \cellcolor{gray!15}\textbf{0.7633} \\
					& 0.10 & 0.7124 & 0.7206 & 0.7164 & 0.7181 & 0.7288 & 0.7286 & 0.7295 & 0.7207 & 0.7158 & \textbf{0.7467} \\
					& \cellcolor{gray!15}0.15 & \cellcolor{gray!15}0.6957 & \cellcolor{gray!15}0.7028 & \cellcolor{gray!15}0.6967 & \cellcolor{gray!15}0.6986 & \cellcolor{gray!15}0.7106 & \cellcolor{gray!15}0.7148 & \cellcolor{gray!15}0.7125 & \cellcolor{gray!15}0.7066 & \cellcolor{gray!15}0.6973 & \cellcolor{gray!15}\textbf{0.7377} \\
					& 0.20 & 0.6888 & 0.6977 & 0.6907 & 0.6916 & 0.7021 & 0.7136 & 0.7017 & 0.7037 & 0.6924 & \textbf{0.7254} \\
					& \cellcolor{gray!15}0.25 & \cellcolor{gray!15}0.6771 & \cellcolor{gray!15}0.6826 & \cellcolor{gray!15}0.6815 & \cellcolor{gray!15}0.6817 & \cellcolor{gray!15}0.6883 & \cellcolor{gray!15}0.7001 & \cellcolor{gray!15}0.6876 & \cellcolor{gray!15}0.6936 & \cellcolor{gray!15}0.6787 & \cellcolor{gray!15}\textbf{0.7007} \\
					\hline
					\multirow{6}{*}{\textbf{Pubmed}} & 0.00 & 0.8588 & 0.8475 & 0.8608 & 0.8601 & 0.8567 & 0.8751 & 0.8502 & 0.8349 & 0.8594 & \textbf{0.8824} \\
					& \cellcolor{gray!15}0.05 & \cellcolor{gray!15}0.8002 & \cellcolor{gray!15}0.7952 & \cellcolor{gray!15}0.8019 & \cellcolor{gray!15}0.8002 & \cellcolor{gray!15}0.7994 & \cellcolor{gray!15}0.8381 & \cellcolor{gray!15}0.7966 & \cellcolor{gray!15}0.8127 & \cellcolor{gray!15}0.7999 & \cellcolor{gray!15}\textbf{0.8478} \\
					& 0.10 & 0.7493 & 0.7436 & 0.7510 & 0.7506 & 0.7508 & \textbf{0.8119} & 0.7519 & 0.7933 & 0.7489 & 0.7969 \\
					& \cellcolor{gray!15}0.15 & \cellcolor{gray!15}0.7007 & \cellcolor{gray!15}0.6920 & \cellcolor{gray!15}0.7027 & \cellcolor{gray!15}0.7016 & \cellcolor{gray!15}0.7043 & \cellcolor{gray!15}\textbf{0.7946} & \cellcolor{gray!15}0.7094 & \cellcolor{gray!15}0.7774 & \cellcolor{gray!15}0.7017 & \cellcolor{gray!15}0.7799 \\
					& 0.20 & 0.6617 & 0.6489 & 0.6638 & 0.6632 & 0.6660 & \textbf{0.7768} & 0.6708 & 0.7648 & 0.6626 & 0.7135 \\
					& \cellcolor{gray!15}0.25 & \cellcolor{gray!15}0.6249 & \cellcolor{gray!15}0.6101 & \cellcolor{gray!15}0.6270 & \cellcolor{gray!15}0.6268 & \cellcolor{gray!15}0.6287 & \cellcolor{gray!15}\textbf{0.7714} & \cellcolor{gray!15}0.6351 & \cellcolor{gray!15}0.7519 & \cellcolor{gray!15}0.6275 & \cellcolor{gray!15}0.6800 \\
					\hline
					\multirow{6}{*}{\shortstack{\textbf{Amazon}\\\textbf{Photo}}} & 0.00 & 0.8672 & 0.9383 & 0.8938 & 0.9287 & 0.9072 & 0.8717 & 0.8599 & 0.5514 & 0.9308 & \textbf{0.9638} \\
					& \cellcolor{gray!15}0.05 & \cellcolor{gray!15}0.7894 & \cellcolor{gray!15}0.8369 & \cellcolor{gray!15}0.8428 & \cellcolor{gray!15}0.8518 & \cellcolor{gray!15}0.8241 & \cellcolor{gray!15}0.7958 & \cellcolor{gray!15}0.7952 & \cellcolor{gray!15}0.4661 & \cellcolor{gray!15}0.8296 & \cellcolor{gray!15}\textbf{0.8587} \\
					& 0.10 & 0.7582 & 0.7819 & 0.8037 & 0.8067 & 0.7996 & 0.7633 & 0.7587 & 0.4054 & 0.7773 & \textbf{0.8203} \\
					& \cellcolor{gray!15}0.15 & \cellcolor{gray!15}0.7081 & \cellcolor{gray!15}0.7458 & \cellcolor{gray!15}0.7708 & \cellcolor{gray!15}0.7789 & \cellcolor{gray!15}0.7534 & \cellcolor{gray!15}0.7441 & \cellcolor{gray!15}0.7153 & \cellcolor{gray!15}0.3849 & \cellcolor{gray!15}0.7406 & \cellcolor{gray!15}\textbf{0.7944} \\
					& 0.20 & 0.6899 & 0.7124 & 0.7372 & 0.7491 & 0.7258 & 0.7079 & 0.7019 & 0.3425 & 0.7049 & \textbf{0.7687} \\
					& \cellcolor{gray!15}0.25 & \cellcolor{gray!15}0.6453 & \cellcolor{gray!15}0.6780 & \cellcolor{gray!15}0.7025 & \cellcolor{gray!15}0.7136 & \cellcolor{gray!15}0.7038 & \cellcolor{gray!15}0.6925 & \cellcolor{gray!15}0.6659 & \cellcolor{gray!15}0.3173 & \cellcolor{gray!15}0.6661 & \cellcolor{gray!15}\textbf{0.7485} \\
					\hline
				\end{tabular}
			}
			\label{tab_a}
		}
		
		\subfloat[Accuracy under Min-Max attack on disassortative graphs]
		{\resizebox{\textwidth}{!}
			{\begin{tabular}{c|c|cccccccccc}
					\hline
					\textbf{Dataset} & \textbf{Ptb Rate} & \textbf{GCN} & \textbf{GAT} & \textbf{GCN-Jaccard} & \textbf{RGCN} & \textbf{GCN-SVD} & \textbf{SimPGCN} & \textbf{ERGCN} & \textbf{ADGCN} & \textbf{GraphReshape} & \textbf{GJDNet} \\
					\hline
					\multirow{6}{*}{\textbf{Cornell}} & 0.00 & 0.3415 & 0.4340 & 0.4347 & 0.3884 & 0.3544 & 0.4687 & 0.4279 & 0.5347 & 0.4095 & \textbf{0.5504} \\
					& \cellcolor{gray!15}0.05 & \cellcolor{gray!15}0.3395 & \cellcolor{gray!15}0.4313 & \cellcolor{gray!15}0.3850 & \cellcolor{gray!15}0.3633 & \cellcolor{gray!15}0.3435 & \cellcolor{gray!15}0.4537 & \cellcolor{gray!15}0.4190 & \cellcolor{gray!15}0.5211 & \cellcolor{gray!15}0.3884 & \cellcolor{gray!15}\textbf{0.5379} \\
					& 0.10 & 0.3367 & 0.4272 & 0.3721 & 0.3524 & 0.3415 & 0.4449 & 0.4034 & 0.5082 & 0.3850 & \textbf{0.5175} \\
					& \cellcolor{gray!15}0.15 & \cellcolor{gray!15}0.3224 & \cellcolor{gray!15}0.4116 & \cellcolor{gray!15}0.3687 & \cellcolor{gray!15}0.3490 & \cellcolor{gray!15}0.3252 & \cellcolor{gray!15}0.4401 & \cellcolor{gray!15}0.3891 & \cellcolor{gray!15}0.4926 & \cellcolor{gray!15}0.3755 & \cellcolor{gray!15}\textbf{0.5036} \\
					& 0.20 & 0.3129 & 0.4020 & 0.3605 & 0.3224 & 0.3170 & 0.4381 & 0.3660 & 0.4748 & 0.3728 & \textbf{0.4820} \\
					& \cellcolor{gray!15}0.25 & \cellcolor{gray!15}0.3007 & \cellcolor{gray!15}0.3837 & \cellcolor{gray!15}0.3490 & \cellcolor{gray!15}0.3116 & \cellcolor{gray!15}0.3109 & \cellcolor{gray!15}0.4204 & \cellcolor{gray!15}0.3476 & \cellcolor{gray!15}0.4565 & \cellcolor{gray!15}0.3694 & \cellcolor{gray!15}\textbf{0.4646} \\
					\hline
					\multirow{6}{*}{\textbf{Texas}} & 0.00 & 0.4644 & 0.5562 & 0.5500 & 0.5521 & 0.4938 & 0.5829 & 0.5397 & 0.5589 & 0.5144 & \textbf{0.6667} \\
					& \cellcolor{gray!15}0.05 & \cellcolor{gray!15}0.4568 & \cellcolor{gray!15}0.5514 & \cellcolor{gray!15}0.5212 & \cellcolor{gray!15}0.5247 & \cellcolor{gray!15}0.4678 & \cellcolor{gray!15}0.5733 & \cellcolor{gray!15}0.5342 & \cellcolor{gray!15}0.5575 & \cellcolor{gray!15}0.5096 & \cellcolor{gray!15}\textbf{0.6491} \\
					& 0.10 & 0.4493 & 0.5473 & 0.4986 & 0.5253 & 0.4541 & 0.5589 & 0.5260 & 0.5500 & 0.4856 & \textbf{0.6328} \\
					& \cellcolor{gray!15}0.15 & \cellcolor{gray!15}0.4322 & \cellcolor{gray!15}0.5438 & \cellcolor{gray!15}0.4801 & \cellcolor{gray!15}0.5116 & \cellcolor{gray!15}0.4445 & \cellcolor{gray!15}0.5466 & \cellcolor{gray!15}0.5151 & \cellcolor{gray!15}0.5438 & \cellcolor{gray!15}0.4842 & \cellcolor{gray!15}\textbf{0.6222} \\
					& 0.20 & 0.4199 & 0.5377 & 0.4637 & 0.5062 & 0.4295 & 0.5418 & 0.5014 & 0.5342 & 0.4747 & \textbf{0.6142} \\
					& \cellcolor{gray!15}0.25 & \cellcolor{gray!15}0.4158 & \cellcolor{gray!15}0.5356 & \cellcolor{gray!15}0.4486 & \cellcolor{gray!15}0.4801 & \cellcolor{gray!15}0.4171 & \cellcolor{gray!15}0.5377 & \cellcolor{gray!15}0.4952 & \cellcolor{gray!15}0.5315 & \cellcolor{gray!15}0.4719 & \cellcolor{gray!15}\textbf{0.6050} \\
					\hline
					\multirow{6}{*}{\textbf{Wisconsin}} & 0.00 & 0.4045 & 0.4886 & 0.4746 & 0.4463 & 0.4507 & 0.6090 & 0.4731 & 0.6114 & 0.4453 & \textbf{0.6138} \\
					& \cellcolor{gray!15}0.05 & \cellcolor{gray!15}0.3960 & \cellcolor{gray!15}0.4711 & \cellcolor{gray!15}0.4667 & \cellcolor{gray!15}0.4408 & \cellcolor{gray!15}0.4423 & \cellcolor{gray!15}0.5900 & \cellcolor{gray!15}0.4537 & \cellcolor{gray!15}0.5910 & \cellcolor{gray!15}0.4378 & \cellcolor{gray!15}\textbf{0.5926} \\
					& 0.10 & 0.3871 & 0.4706 & 0.4602 & 0.4348 & 0.4403 & 0.5811 & 0.4512 & 0.5861 & 0.4239 & \textbf{0.5876} \\
					& \cellcolor{gray!15}0.15 & \cellcolor{gray!15}0.3811 & \cellcolor{gray!15}0.4552 & \cellcolor{gray!15}0.4443 & \cellcolor{gray!15}0.4333 & \cellcolor{gray!15}0.4378 & \cellcolor{gray!15}0.5781 & \cellcolor{gray!15}0.4428 & \cellcolor{gray!15}0.5801 & \cellcolor{gray!15}0.4214 & \cellcolor{gray!15}\textbf{0.5828} \\
					& 0.20 & 0.3751 & 0.4473 & 0.4368 & 0.4274 & 0.4259 & 0.5657 & 0.4313 & 0.5766 & 0.4169 & \textbf{0.5793} \\
					& \cellcolor{gray!15}0.25 & \cellcolor{gray!15}0.3652 & \cellcolor{gray!15}0.4065 & \cellcolor{gray!15}0.4109 & \cellcolor{gray!15}0.4134 & \cellcolor{gray!15}0.4199 & \cellcolor{gray!15}0.5617 & \cellcolor{gray!15}0.4269 & \cellcolor{gray!15}0.5612 & \cellcolor{gray!15}0.4060 & \cellcolor{gray!15}\textbf{0.5724} \\
					\hline
					\multirow{6}{*}{\textbf{Actor}} & 0.00 & 0.2614 & 0.2804 & 0.2823 & 0.2676 & 0.2762 & 0.2787 & 0.2625 & 0.3104 & 0.2671 & \textbf{0.3337} \\
					& \cellcolor{gray!15}0.05 & \cellcolor{gray!15}0.2578 & \cellcolor{gray!15}0.2745 & \cellcolor{gray!15}0.2660 & \cellcolor{gray!15}0.2622 & \cellcolor{gray!15}0.2639 & \cellcolor{gray!15}0.2739 & \cellcolor{gray!15}0.2605 & \cellcolor{gray!15}0.3076 & \cellcolor{gray!15}0.2604 & \cellcolor{gray!15}\textbf{0.3127} \\
					& 0.10 & 0.2564 & 0.2681 & 0.2635 & 0.2605 & 0.2602 & 0.2704 & 0.2599 & 0.2977 & 0.2582 & \textbf{0.3030} \\
					& \cellcolor{gray!15}0.15 & \cellcolor{gray!15}0.2547 & \cellcolor{gray!15}0.2657 & \cellcolor{gray!15}0.2617 & \cellcolor{gray!15}0.2588 & \cellcolor{gray!15}0.2598 & \cellcolor{gray!15}0.2660 & \cellcolor{gray!15}0.2592 & \cellcolor{gray!15}0.2916 & \cellcolor{gray!15}0.2579 & \cellcolor{gray!15}\textbf{0.2952} \\
					& 0.20 & 0.2516 & 0.2638 & 0.2604 & 0.2527 & 0.2553 & 0.2630 & 0.2589 & 0.2853 & 0.2549 & \textbf{0.2891} \\
					& \cellcolor{gray!15}0.25 & \cellcolor{gray!15}0.2496 & \cellcolor{gray!15}0.2611 & \cellcolor{gray!15}0.2595 & \cellcolor{gray!15}0.2545 & \cellcolor{gray!15}0.2535 & \cellcolor{gray!15}0.2609 & \cellcolor{gray!15}0.2566 & \cellcolor{gray!15}0.2839 & \cellcolor{gray!15}0.2523 & \cellcolor{gray!15}\textbf{0.2860} \\
					\hline
			\end{tabular}}
		}
		\label{tab_b}}
\end{table*}

\subsection{Robustness Against Adversarial Attacks (RQ1)}
\label{sec:Robustness_Against_Adversarial_Attacks}
In this subsection, we evaluate robustness under structural perturbations. Table \ref{tab2_minmax} reports the node classification accuracy under Min-Max attacks with perturbation rates from 0.00 to 0.25. On assortative graphs, GJDNet achieves the best performance on Cora, Citeseer, and Amazon Photo across the full perturbation range. On Pubmed at higher rates ($\geq$0.10), SimPGCN performs better, likely because its feature-similarity reconstruction reduces sensitivity to perturbed edges. On disassortative graphs, GJDNet consistently outperforms all baselines with a graceful degradation trend as perturbations increase. These results indicate that GJDNet achieves strong robustness under global Min-Max attacks across both assortative and disassortative graphs.

\begin{figure*}[!t]
	\centering
	\includegraphics[width=\textwidth,height=100pt]{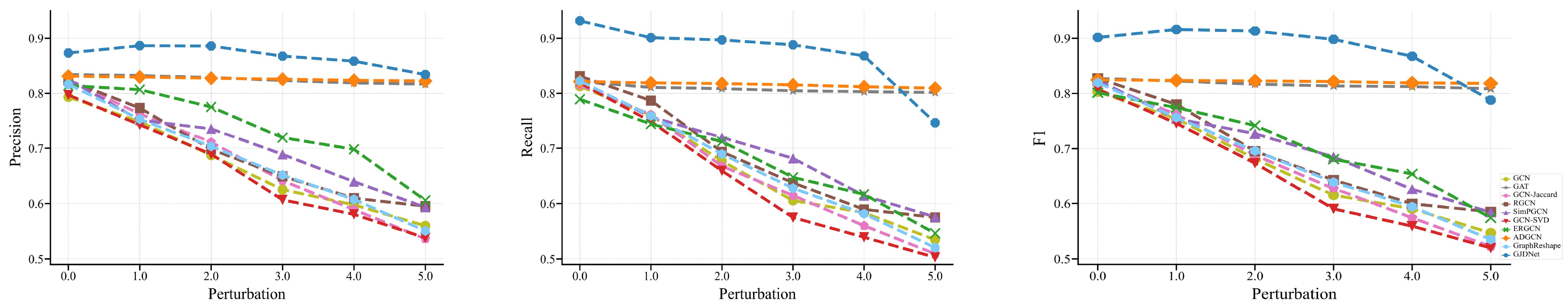}
	\caption{Precision, Recall, and F1 score of node classification methods on Cora under Nettack. The horizontal axis represents the number of perturbations applied to each target node (0--5), while the vertical axis denotes the corresponding evaluation metric value (Precision, Recall, or F1 score).}
	\label{fig5_precision}
\end{figure*}
\begin{figure*}[!t]
	\centering
	\includegraphics[width=\textwidth,height=180pt]{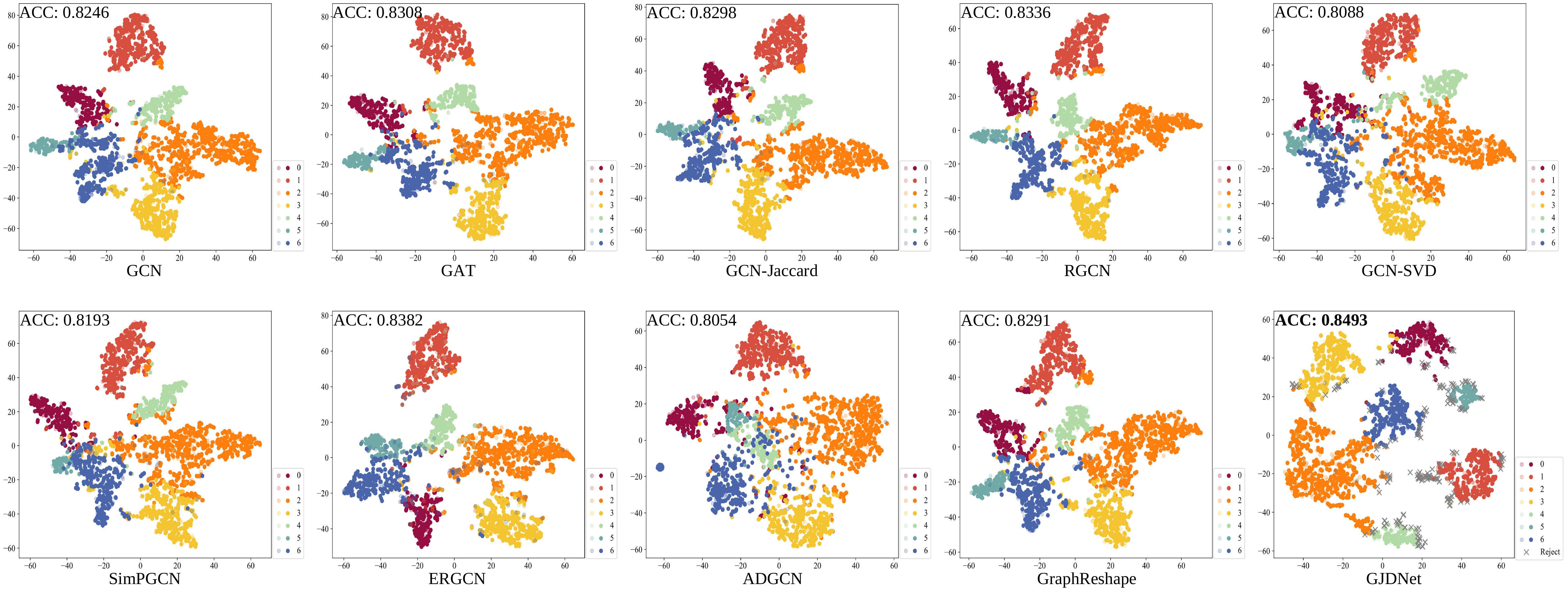}
	\caption{Visualization of decision boundaries of Cora under Random Attack with a 5\% perturbation rate. Training and validation samples are shown with lighter colors, while testing samples are shown with darker colors. In GJDNet, rejected samples are explicitly visualized as gray crosses, i.e., testing samples that fall into the rejection region of the SDB.}
	\label{fig6_vis_rand_cora}
\end{figure*}
\begin{figure*}[!t]
	\centering
	\subfloat[Neighbor-feature heatmaps of the target node]{
		\includegraphics[width=\textwidth]{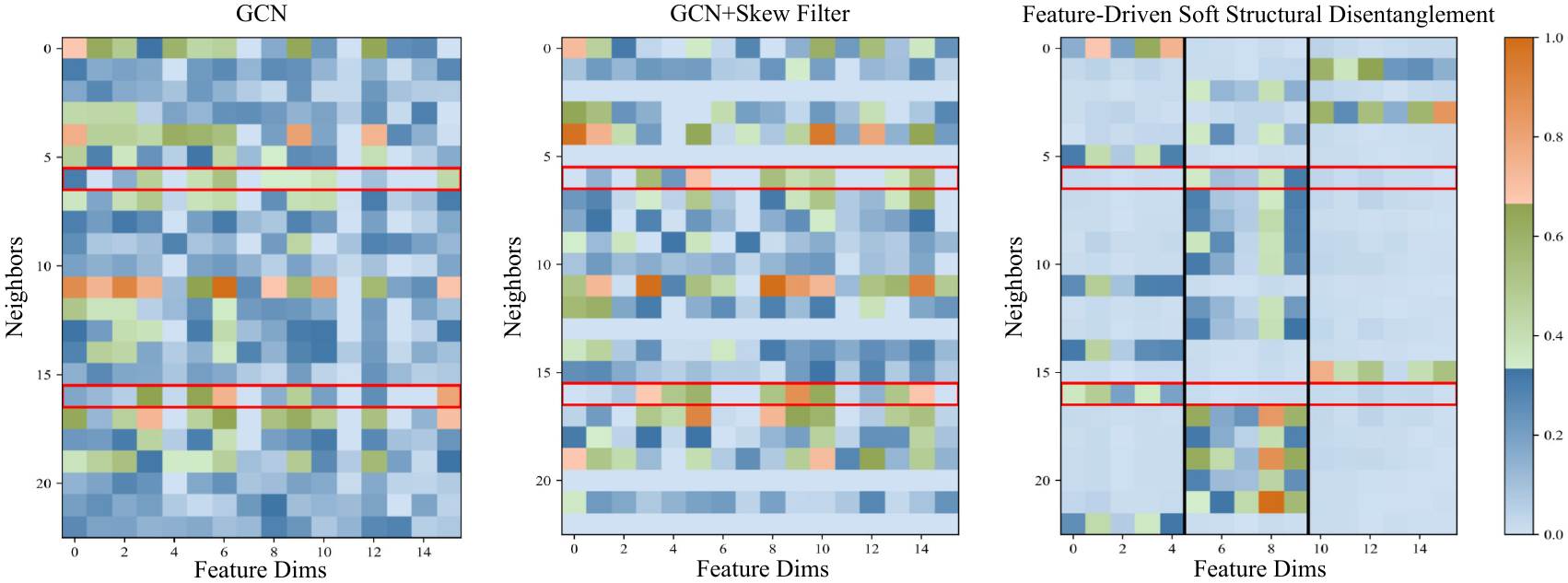}
		\label{fig7_disentangle_a}
	}
	
	\subfloat[Embedding dimensions affected by adversarial perturbations]{
		\includegraphics[width=\textwidth]{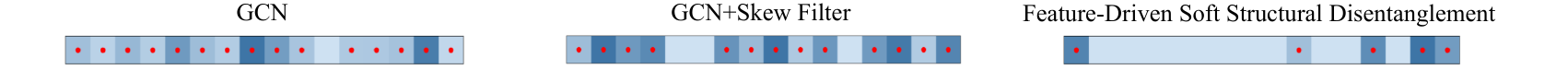}
		\label{fig7_disentangle_b}
	}
	
	\caption{Visualization of adversarial perturbation effects on the Cora dataset under the Nettack (2.0 perturbation number)  across three architectures: GCN, GCN+Skew Filter, and Feature-driven Soft Structural Disentanglement. Red boxes highlight adversarially perturbed neighbors, while red markers indicate embedding dimensions affected by the perturbation after aggregation.}
	\label{fig7_disentangle}
\end{figure*}

\begin{figure*}[!t]
	\centering
	\includegraphics[width=\textwidth,height=240pt]{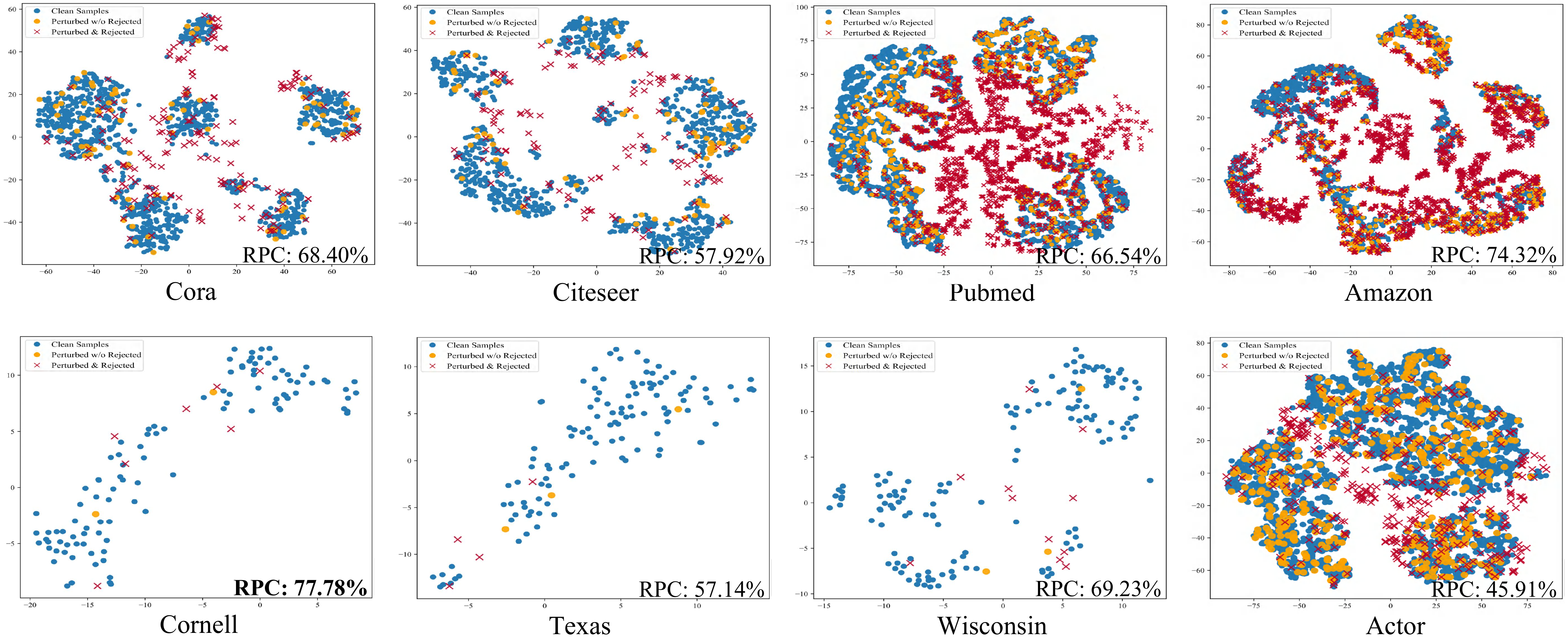}
	\caption{Rejection-Perturbation Coverage (RPC) visualization under the Min-Max attack (5\% perturbation). Clean samples are shown in blue, perturbed but unrejected samples are visualized as orange circles, while adversarial samples rejected by the SDB decision rule are highlighted as red crosses.}
	\label{fig8_sdb_reject_minmax}
\end{figure*}

Under Nettack on Cora (Fig. \ref{fig5_precision}), all baselines degrade as perturbations increase. Methods relying on static neighborhood aggregation (e.g., GCN and RGCN) or similarity-based filtering (e.g., GCN-Jaccard) show sharp drops beyond two perturbations, indicating sensitivity to localized noise. In contrast, GJDNet remains stable as perturbations increase.

To intuitively illustrate the robustness and geometry-aware discriminative properties of the learned node embeddings in the decision space, Fig. \ref{fig6_vis_rand_cora} visualizes t-SNE embeddings on Cora under Random Attack (5\%). Conventional models exhibit overlapping clusters and weak separation, indicating reduced class margins under perturbations. In contrast, GJDNet not only forms compact intra-class clusters with clearly separated inter-class margins, but also explicitly rejects a subset of perturbed test samples located outside the SDB regions.

\begin{table*}[!t]
	\caption{Ablation results on six datasets under the Min-Max attack with 5\% perturbation rate.}
	\centering
	{
		\fontsize{7pt}{6pt}\selectfont
		\renewcommand{\arraystretch}{1.5}
		\setlength{\tabcolsep}{3pt}
		\begin{tabularx}{\textwidth}{>{\centering\arraybackslash}p{0.18\textwidth}
				*{8}{>{\centering\arraybackslash}X}}
			\hline
			\textbf{Ablation} & \textbf{Cora} & \textbf{Citeseer} & \textbf{Pubmed} & \textbf{Amazon} & \textbf{Cornell} & \textbf{Texas} & \textbf{Wisconsin} & \textbf{Actor} \\
			\hline
			\textbf{w/o Skew Filter} & 0.8345 & 0.7506 & 0.8455 & 0.8540 & 0.4082 & 0.5354 & 0.5124 & 0.2502 \\
			\rowcolor{gray!15}
			\textbf{w/o Incoming Attention} & 0.8434 & 0.7478 & 0.8351 & 0.8479 & 0.5170 & 0.6048 & 0.5204 & 0.2598 \\			
			\textbf{w/o Outgoing Attention} & 0.8254 & 0.7612 & 0.8404 & 0.8457 & 0.4490 & 0.6063 & 0.5821 & 0.3057 \\
			\rowcolor{gray!15}
			\textbf{w/o Dual Attention} & 0.8398 & 0.7251 & 0.8276 & 0.8075 & 0.5109 & 0.6033 & 0.5473 & 0.2557 \\
			\textbf{w/o Routing} & 0.8460 & 0.7535 & 0.8421 & 0.8477 & 0.5299 & 0.5758 & 0.5417 & 0.2838 \\
			\rowcolor{gray!15}
			\textbf{w/o Disentanglement} & 0.8389 & 0.7250 & 0.8268 & 0.8050 & 0.3380 & 0.5407 & 0.4508 & 0.2497 \\
			\textbf{w/o SDB-$\mathcal{L}_{\mathrm{intra}}$} & 0.8367 & 0.7398 & 0.8275 & 0.8414 & 0.5267 & 0.5968 & 0.5027 & 0.2823 \\
			\rowcolor{gray!15}
			\textbf{w/o SDB-$\mathcal{L}_{\mathrm{inter}}$} & 0.8362 & 0.7339 & 0.8204 & 0.8496 & 0.5301 & 0.6015 & 0.5323 & 0.2935 \\
			\textbf{w/o SDB Loss} & 0.8288 & 0.7293 & 0.8199 & 0.8388 & 0.5221 & 0.5859 & 0.4950 & 0.2856 \\
			\rowcolor{gray!15}
			\textbf{w/o Classification} & 0.8209 & 0.7275 & 0.8044 & 0.8344 & 0.4658 & 0.5890 & 0.4876 & 0.2821 \\
			\textbf{GJDNet} & \textbf{0.8547} & \textbf{0.7633} & \textbf{0.8478} & \textbf{0.8587} & \textbf{0.5379} & \textbf{0.6491} & \textbf{0.5926} & \textbf{0.3127} \\
			\hline
		\end{tabularx}
	}
	\label{ablation}
\end{table*}

\subsection{Mechanism Analysis of GJDNet (RQ2)}
\label{sec:Mechanism_Analysis_of_GJDNet}
\textbf{Feature-driven Soft Structural Disentanglement Analysis.}
To elucidate how feature-driven soft structural disentanglement enhances robustness, we analyze the effect of adversarial perturbations in Fig. \ref{fig7_disentangle}. On Cora, Nettack introduces mismatched neighbors through connectivity inversion, and GCN propagates these perturbations via uniform aggregation, resulting in diffuse perturbation responses in both neighbor–feature heatmaps and node embeddings. Skewness-aware filtering reduces the magnitude of these responses by pruning abnormal neighbors; however, since aggregation remains unchanged, perturbations can still spread across dimensions. In contrast, soft structural disentanglement decomposes representations into multiple subspaces and routes low-consistency neighbors into constrained factors. As shown in the third column of Fig. \ref{fig7_disentangle}, perturbations remain confined to a limited number of subspaces and embedding dimensions, while others are largely unaffected. This subspace isolation preserves feature-consistent representations and enhances robustness.

\textbf{SDB-Guided Classification Mechanism Analysis.}
Fig. \ref{fig8_sdb_reject_minmax} visualizes the Rejection-Perturbation Coverage (RPC) under Min-Max attack (5\%). Across all datasets, clean samples generally form relatively compact clusters in the embedding space, while many perturbed samples drift away from their class centers and fall outside the corresponding spherical regions, leading to rejection by the SDB decision mechanism. A small subset of perturbed samples lies sufficiently close to the class centers to remain within the corresponding spheres, indicating that SDB maintains tolerance to minor perturbations.

From a quantitative perspective, this rejection behavior is summarized by the RPC metric, defined as the proportion of perturbed test samples that are rejected by the SDB decision rule. As shown in Fig. \ref{fig8_sdb_reject_minmax}, RPC values range from 45.91\% to 77.78\% across datasets, with higher rejection coverage on Cornell and Amazon, while lower coverage appears on Actor and Texas. These results show that the geometry-aware SDB formulation effectively filters adversarial perturbations, yielding a robust and effective decision mechanism. The relatively lower RPC on Actor may be attributed to its weak connectivity regimes and complex feature distribution, which makes adversarial perturbations harder to separate from clean representations. By embedding class-discriminative semantics into explicit spherical regions, SDB secures clean samples within high-density clusters while systematically excluding adversarially perturbed representations, thereby enabling stable and reliable decision under adversarial perturbations.

\subsection{Ablation Studies (RQ3)}
\label{sec:Ablation_Studies}
To evaluate the contribution of each major component in GJDNet, a comprehensive ablation study is conducted under the Min-Max attack with a 5\% perturbation rate, as summarized in Table \ref{ablation}. For the disentanglement module, removing the skewness-aware filter causes noticeable degradation (e.g., Wisconsin from 0.5926 to 0.5124), highlighting the importance of modeling asymmetric similarity distributions. Directional attention is also essential, as eliminating either direction reduces performance across datasets. Disabling routing leads to larger drops on disassortative graphs (e.g., Wisconsin from 0.5926 to 0.5417), indicating the role of adaptive cross-subspace weighting under structural perturbations. Removing multi-subspace disentanglement further weakens robustness.

For the spherical decision boundary module, removing SDB-related constraints (intra-class, inter-class, or full regularization) consistently reduces robustness. Eliminating the full SDB regularizer degrades accuracy (e.g., Wisconsin from 0.5926 to 0.4950), and removing the sphere-based classification rule further lowers performance (e.g., Wisconsin from 0.5926 to 0.4876). These results demonstrate that soft multi-subspace disentanglement and spherical decision-space modeling provide complementary robustness gains.

\subsection{Hyperparameter Analysis (RQ4)}
\label{sec:Hyperparameter_Analysis}
\begin{figure}[htbp]
	\centering
	\includegraphics[width=\linewidth,height=110pt]{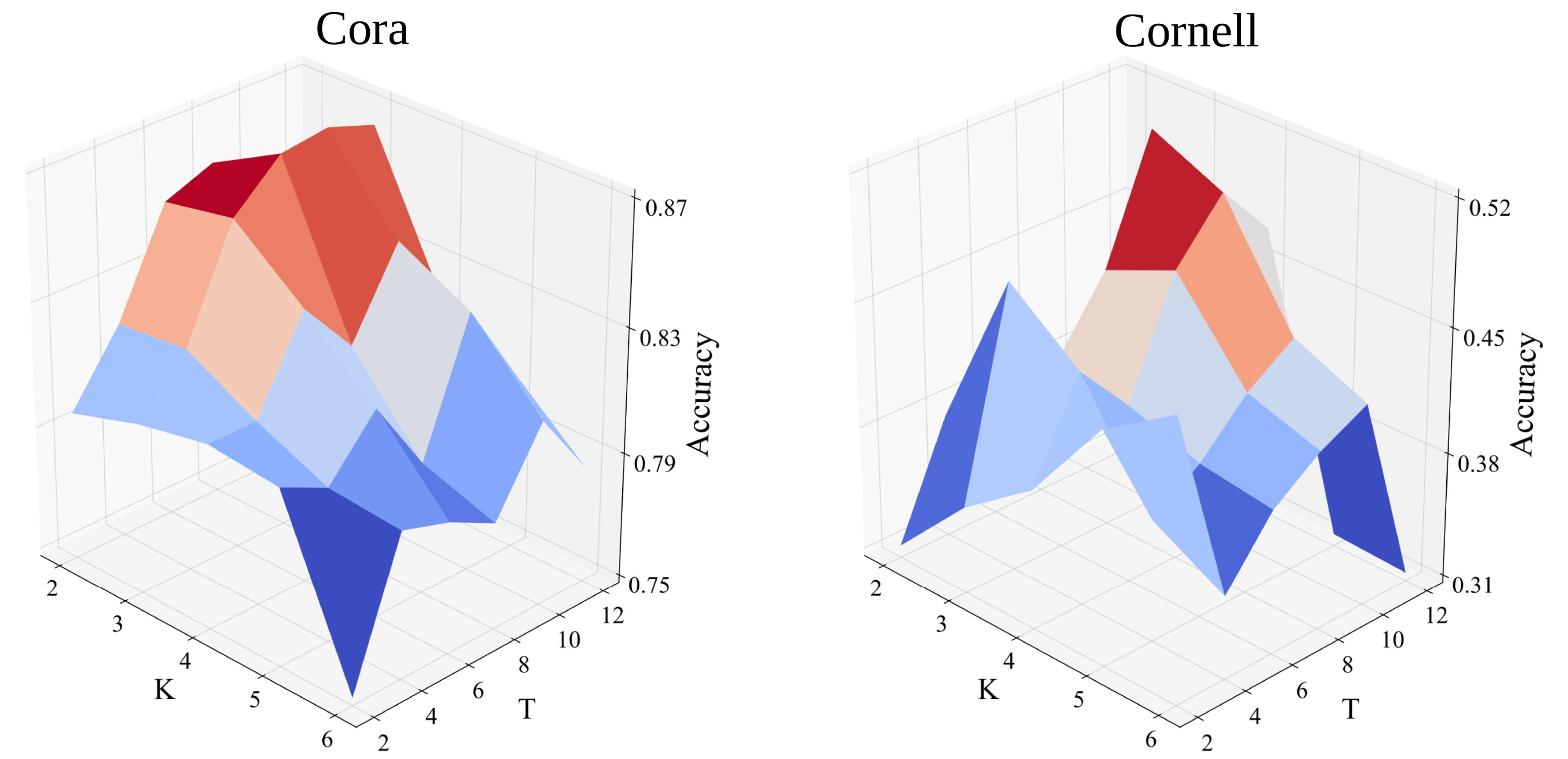}
	\caption{Hyperparameter sensitivity of the soft structural disentanglement module with respect to the number of subspaces $K$ and the number of iterations $T$ on the Cora and Cornell datasets.}
	\label{fig9_parameter_kt}
\end{figure}

This section investigates the sensitivity of GJDNet to key hyperparameters across its three major components, including the soft structural disentanglement module, the skewness-aware neighbor filter, and the spherical decision boundary regularization.
Fig. \ref{fig9_parameter_kt} shows accuracy over different numbers of subspaces $K$ and iterations $T$. On both Cora and Cornell, the accuracy forms a clear nonlinear landscape over the $K$--$T$ plane. Increasing either $K$ or $T$ from small values improves performance, as richer subspace disentanglement and deeper iterative refinement allow more expressive structural modeling. When $K$ or $T$ becomes excessively large, however, the performance declines due to over-factorization or repeated propagation, which may accumulate noise and introduce redundant message passing. Optimal performance is achieved at moderate settings (e.g., $K\in[3,5]$, $T\in[6,10]$).

\begin{figure}[!h]
	\centering
	\includegraphics[width=\linewidth,height=120pt]{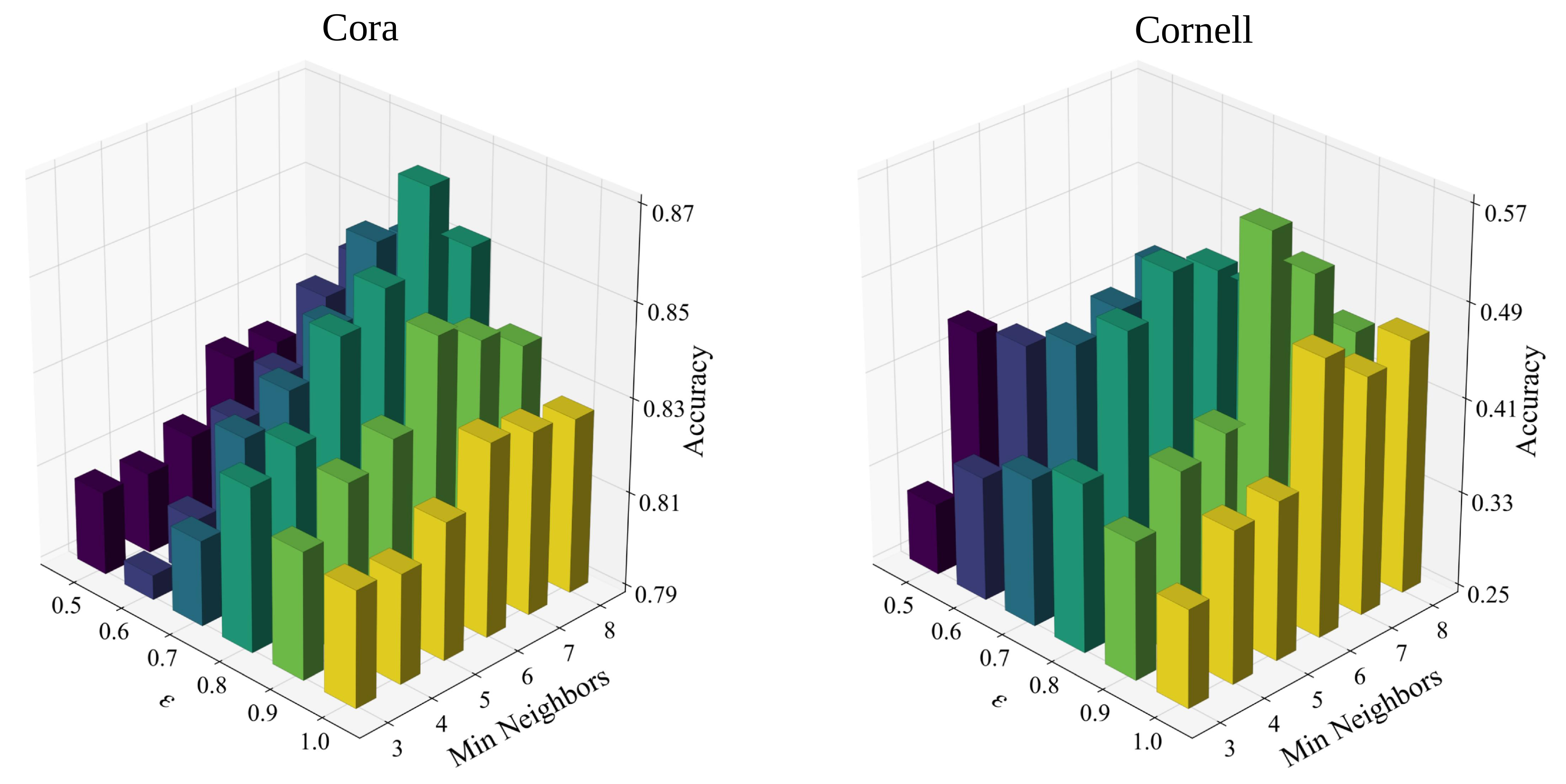}
	\caption{
		Hyperparameter sensitivity of the skewness-aware neighbor filter.
	}
	\label{fig10_parameter_epsilonminn}
\end{figure}

Fig. \ref{fig10_parameter_epsilonminn} visualizes the joint performance surface of the skewness-aware neighbor filter over $(\epsilon,\,{min\_neighbors})$ on Cora and Cornell. The 3D landscape reveals a distinct local peak, highlighting the interaction between $\epsilon$ and ${min\_neighbors}$. Specifically, the highest accuracy is achieved at $\epsilon=0.8$, ${min\_neighbors}=7$ on Cora and at $\epsilon=0.9$, ${min\_neighbors}=6$ on Cornell. Moving away from this peak in either direction—toward overly small or overly large $\epsilon$, or toward extreme ${min\_neighbors}$ values—leads to visible performance decline. These results indicate that optimal robustness arises from a balance between selectivity and neighborhood preservation. The slight difference in optimal settings reflects their structural heterogeneity.

\begin{figure}[!h]
	\centering
	\includegraphics[width=0.8\linewidth,height=110pt]{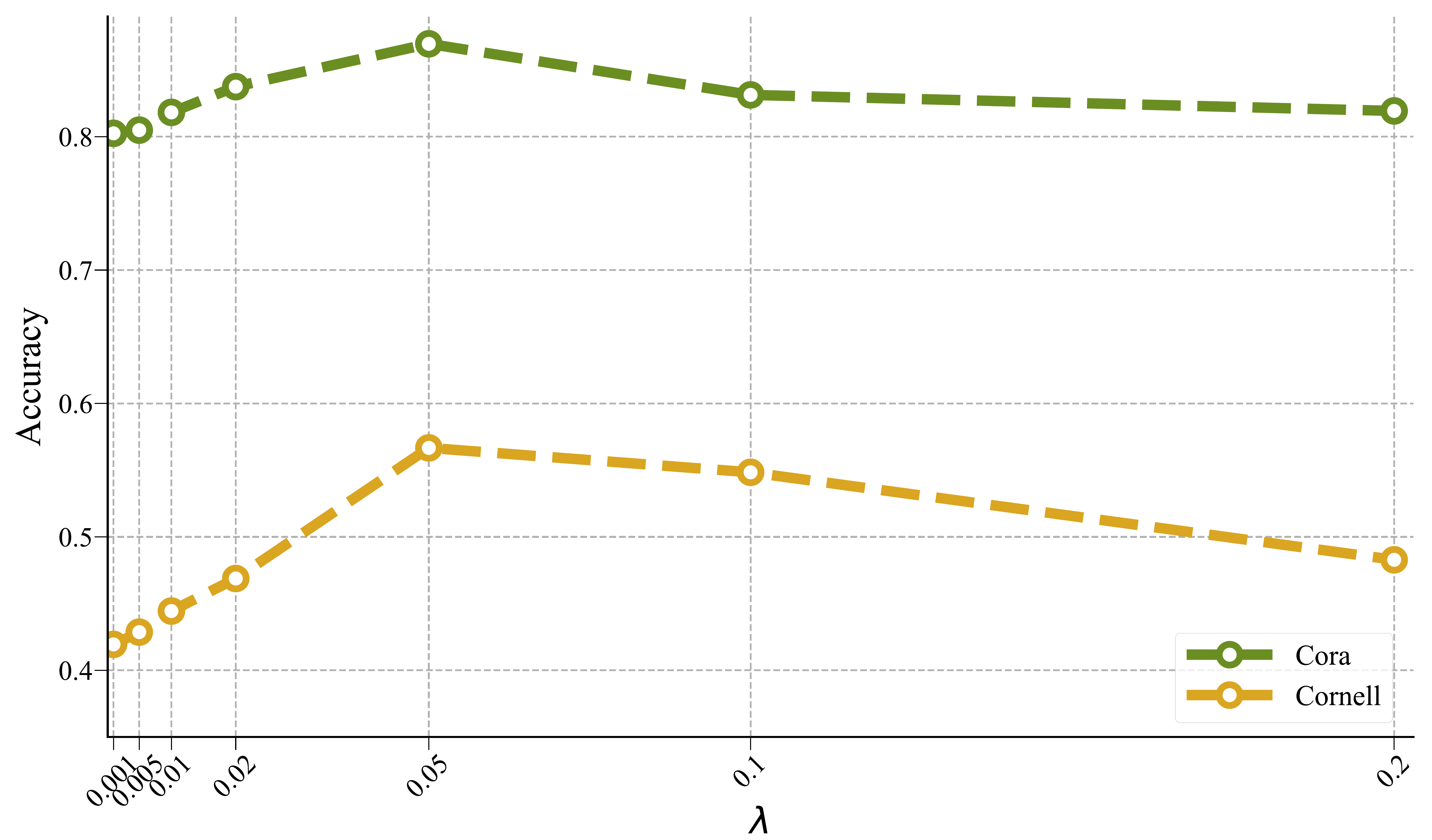}
	\caption{
		Hyperparameter sensitivity of the spherical decision boundary module with respect to $\lambda$.
	}
	\label{fig11_parameter_lambda}
\end{figure}

At the decision level, Fig. \ref{fig11_parameter_lambda} shows the sensitivity of model performance to the SDB regularization coefficient $\lambda$. Both Cora and Cornell exhibit a clear unimodal trend: accuracy increases as $\lambda$ grows from very small values (0.001–0.02) to a moderate level, and then decreases when $\lambda$ becomes large (0.1–0.2), as overly strong geometric constraints limit the expressive capacity of class-wise representations and over-constrain the representation geometry. The peak appears at $\lambda=0.05$ on both datasets (0.8696 on Cora and 0.5667 on Cornell). This indicates that moderate regularization effectively balances geometric compactness and representation flexibility.

\subsection{Complexity and Time Analysis (RQ5)}
\label{sec:Complexity_and_Time_Analysis}
Let $N$ and $E$ denote the numbers of nodes and edges, respectively, and $\mathbf{X}\in\mathbb{R}^{N\times F}$ be the input feature matrix. The model considers $K$ semantic subspaces with embedding dimension $d$. The subspace projection applies $K$ linear mappings, incurring $\mathcal{O}(KNFd)$ cost. Within each iteration, cosine similarity computation over graph edges requires $\mathcal{O}(KEd)$, and skewness-aware filtering performs neighborhood median statistics with an upper bound of $\mathcal{O}(KE\log\Delta)$, where $\Delta$ is the maximum node degree. The dual-directional attention aggregation runs for $T$ sparse message-passing iterations, contributing $\mathcal{O}(TKEd)$. 
Cross-subspace routing and aggregation introduce an additional $\mathcal{O}(KEd)$ cost. The routing term is of the same order and is absorbed into the main complexity. The spherical decision boundary module is edge-independent and requires $\mathcal{O}(Nd)$ for centroid computation and distance evaluation (with the number of classes treated as constant). Overall, the total complexity is $\mathcal{O}\bigl(KNFd + TKE(d + \log\Delta)\bigr)$, which is near-linear in the graph size, up to a $\log\Delta$ factor.

\begin{figure}[htbp]
	\centering
	\includegraphics[width=0.9 \linewidth]{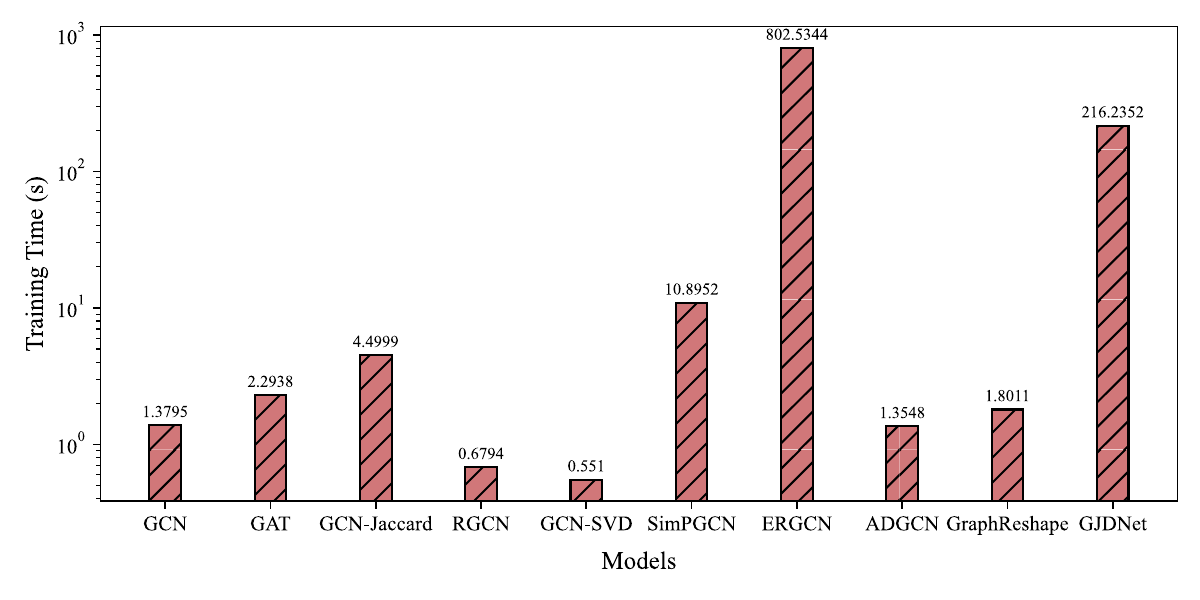}
	\caption{
		Running time comparison of different GNNs.
	}
	\label{fig12_time}
\end{figure}

Empirical running times are reported in Fig. \ref{fig12_time}. Models such as GCN, GAT, and GCN-SVD are the fastest due to their simple message-passing mechanisms. Methods involving structural refinement or similarity modeling incur additional overhead, while ERGCN is the most expensive owing to its iterative optimization procedure. GJDNet remains slower than standard GNNs but substantially more efficient than ERGCN, demonstrating that GJDNet achieves strong robustness with acceptable computational cost.

\section{Conclusion and Discussion}
This paper proposes GJDNet to enhance adversarial robustness in graph learning by jointly improving representation learning and decision-space modeling through feature-driven soft structural disentanglement with skewness-aware neighbor filtering and geometry-constrained spherical decision boundaries. In addition, analysis of adversarial perturbations indicates that such attacks manifest as connectivity pattern inversions that induce structure--feature mismatches and disrupt neighborhood aggregation. The results indicate that adversarial robustness fundamentally depends on the joint disentanglement of node representations and decision spaces, providing a more principled perspective on robust graph learning.

Despite its effectiveness, GJDNet is limited by a fixed number of disentangled subspaces and the lack of explicit modeling of higher-order structural patterns beyond feature-driven neighborhood relations. Future work will focus on learning adaptive disentanglement structures, incorporating explicit structural pattern modeling, and extending the framework to more complex attack settings and dynamic graph scenarios.

\bibliographystyle{IEEEtran}
\bibliography{GJDNet}

\begin{IEEEbiography}[{\includegraphics[width=1in,height=1.25in,clip,keepaspectratio]{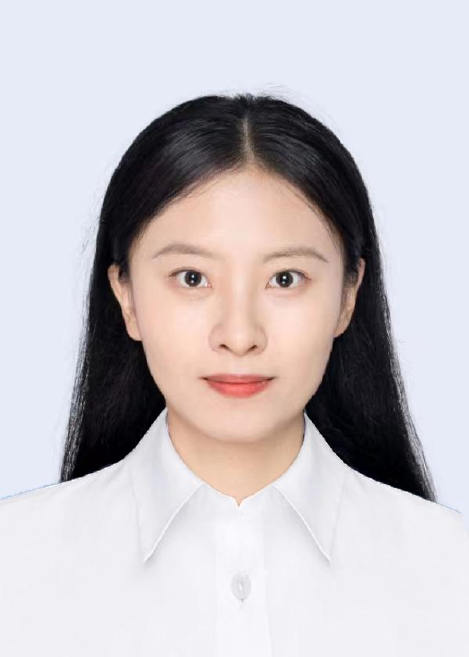}}]
	{Canyixing Cui} received her M.S. degree from Chongqing University of Posts and Telecommunications, Chongqing, China. She is currently pursuing the Ph.D. degree at the School of Computer Science and Technology, Chongqing University of Posts and Telecommunications, Chongqing, China. Her research interests include graph neural networks, adversarial attacks, and robust model design.
\end{IEEEbiography}

\begin{IEEEbiography}[{\includegraphics[width=1in,height=1.25in,clip,keepaspectratio]{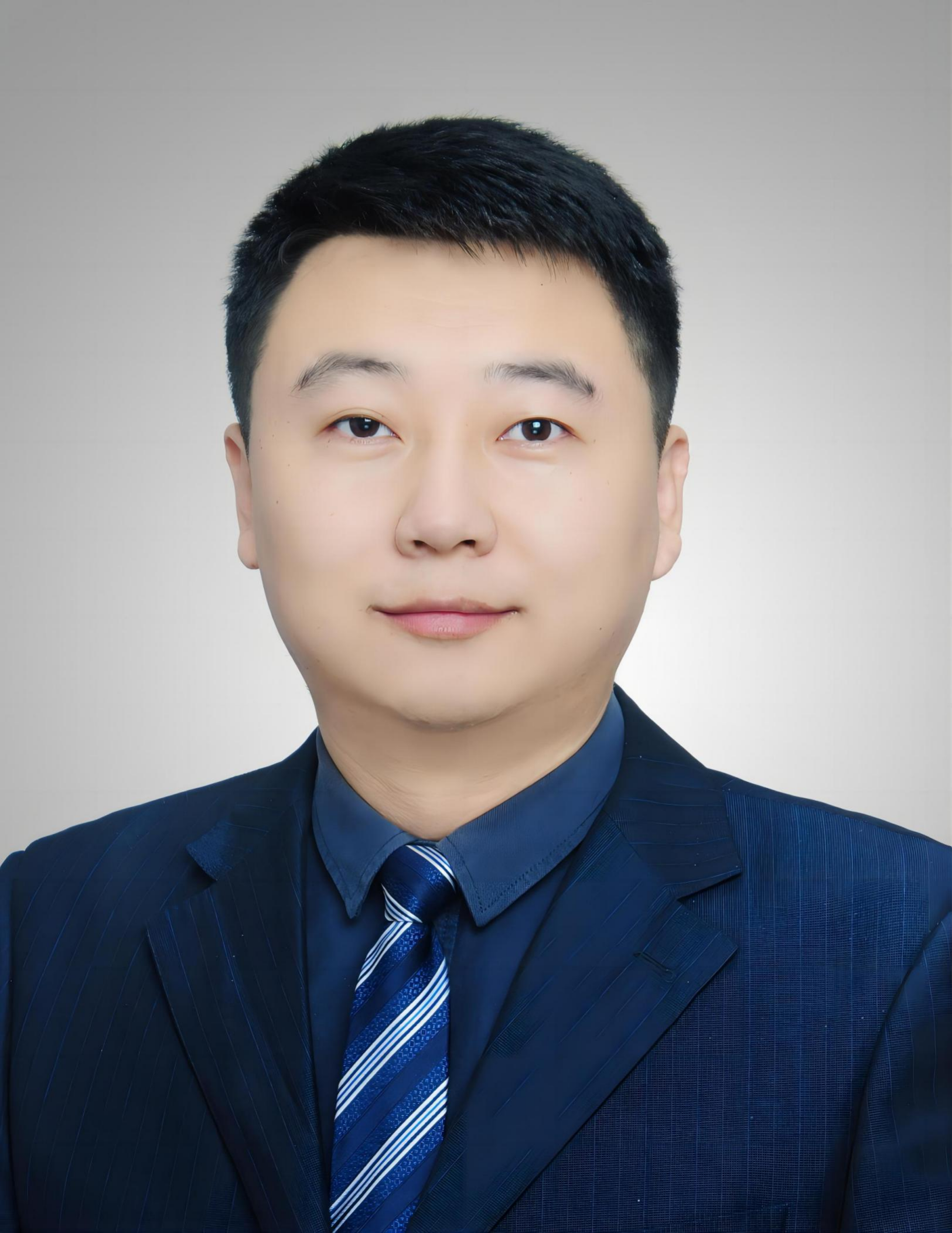}}]
	{Tao Wu} (Member, IEEE) received the Ph.D. degree from the University of Electronic Science and Technology of China in 2017. He is currently a Professor and the Head of the Department of Cybersecurity, Chongqing University of Posts and Telecommunications, China. He previously worked at Tencent Technology Co., Ltd., and currently serves as the Executive Deputy Director of the Chongqing Network and Information Security Technology Engineering Laboratory. He has published over 60 papers in high-impact journals and conferences, including IEEE TIE, IEEE TCSVT, IEEE TBD, IEEE TCSS, and PR. His research interests include graph neural networks, graph foundation models, AI security, and adversarial attack and defense. He has served on the Program Committees of Complex Networks 2023, CSE 2024, and CAAI BDSC 2023–2026.
\end{IEEEbiography}

\begin{IEEEbiography}[{\includegraphics[width=1in,height=1.25in,clip,keepaspectratio]{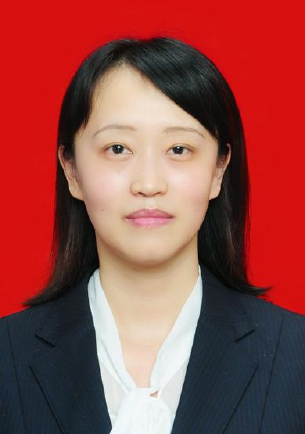}}]
	{Xingping Xian} received her Ph.D. degree from Chongqing University of Posts and Telecommunications, Chongqing, China. She is currently an Associate Professor of the School of Cyber Security and Information Law of the same university. She has published more than ten papers in international journals and conferences. Her research interests include graph data mining, data privacy protection, and intelligent algorithm security.
\end{IEEEbiography}

\begin{IEEEbiography}[{\includegraphics[width=1in,height=1.25in,clip,keepaspectratio]{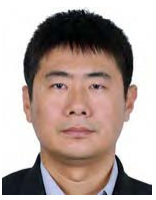}}]
	{Xiao-Ke Xu} (Member, IEEE) received the Ph.D. degree in engineering from the College of Information and Communication Engineering, Dalian Maritime University, Dalian, China, in 2008. He is currently a Professor with the School of Journalism and Communication, Beijing Normal University, Beijing, China. He was a Postdoctoral Fellow with The Hong Kong Polytechnic University, Hong Kong, China, and a Visiting Scholar with the City University of Hong Kong, Hong Kong, China. His research interests include complex network analysis, community detection, and data mining in social networks.
\end{IEEEbiography}

\begin{IEEEbiography}[{\includegraphics[width=1in,height=1.25in,clip,keepaspectratio]{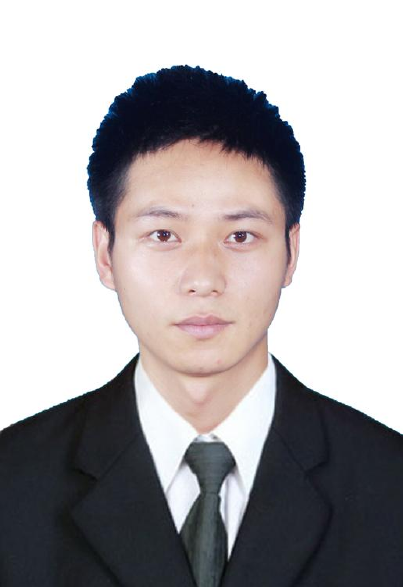}}]
	{Mao Wang} received his M.S. degree from Chongqing University, Chongqing, China, and is currently a Ph.D. candidate of the School of Computer Science and Technology, Chongqing University of Posts and Telecommunications.  His research interests cover a variety of different topics including graph pattern mining, robust graph neural networks, and interpretable machine learning.
\end{IEEEbiography}

\begin{IEEEbiography}[{\includegraphics[width=1in,height=1.25in,clip,keepaspectratio]{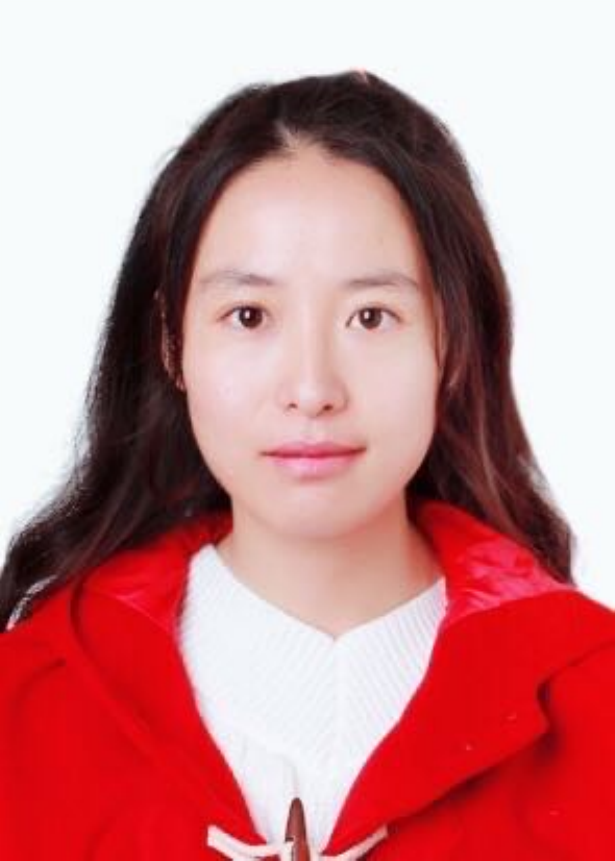}}]
	{Weina Niu} received the bachelor's degree in software engineering from Shenyang Normal University in 2011, and the Ph.D. degree in computer software and theory from the University of Electronic Science and Technology of China in 2018. She is currently a researcher with the School of Computer Science and Engineering, University of Electronic Science and Technology of China. Her research interests include malware analysis, network attack detection, and data security.
\end{IEEEbiography}

\vfill

\end{document}